\documentclass[letterpaper]{article} 
\usepackage[draft]{draft2026}  

\usepackage{times}  
\usepackage{helvet}  
\usepackage{courier}  
\usepackage[hyphens]{url}  
\usepackage{graphicx} 
\usepackage{natbib}  
\usepackage{caption} 
\usepackage{algorithm}  
\usepackage{algorithmic}
\usepackage[table]{xcolor}
\usepackage{multicol}
\usepackage{arydshln}
\usepackage{multirow}
\usepackage{makecell}
\usepackage{enumitem}
\usepackage{booktabs}
\usepackage{tcolorbox}
\usepackage[table]{xcolor}
\usepackage[utf8]{inputenc}
\usepackage[T1]{fontenc} 

\usepackage[framemethod=tikz]{mdframed}

\usepackage{thmtools} 
\usepackage{thm-restate}
\usepackage{lipsum}

\usepackage{fancyhdr,amsmath,amssymb}
\usepackage[ruled,vlined,linesnumbered,algo2e]{algorithm2e}

\SetInd{16pt}{0pt}
\SetNlSty{textbf}{}{\hspace{-8pt}}

\usepackage{newfloat}
\usepackage{listings}
\usepackage{tikz}
\usepackage{etoolbox}
\usepackage{needspace}

\newrobustcmd*\blacka[1]{\tikz[baseline=(char.base)]{
            \node[shape=circle,draw=black,inner sep=1pt,fill=white,text=black,minimum size=1em] (char) {\textsf{\small 1}};}}
\newrobustcmd*\blackb[1]{\tikz[baseline=(char.base)]{
            \node[shape=circle,draw=black,inner sep=1pt,fill=white,text=black,minimum size=1em] (char) {\textsf{\small 2}};}}

\newrobustcmd*\blackc[1]{\tikz[baseline=(char.base)]{
            \node[shape=circle,draw=black,inner sep=1pt,fill=white,text=black,minimum size=1em] (char) {\textsf{\small 3}};}}
\usepackage{microtype}
\definecolor{shadecolor}{rgb}{0.97,0.97,0.97}  
\usepackage{framed} 

\DeclareCaptionStyle{ruled}{labelfont=normalfont,labelsep=colon,strut=off} 
\floatstyle{ruled}
\newfloat{listing}{tb}{lst}{}
\floatname{listing}{Listing}

\title{{\textbf{ReaLM}: Reflection-Enhanced Autonomous Reasoning with Small Language Models}}
\author{
    Yuanfeng Xu\textsuperscript{\rm 1,2}\equalcontrib,
    Zehui Dai\textsuperscript{\rm 1}\equalcontrib \footnotemark[2],
    Jian Liang\textsuperscript{\rm 3},
    Jiapeng Guan\textsuperscript{\rm 4},\\ 
    Guangrun Wang\textsuperscript{\rm 2} \footnotemark[2],
    Liang Lin\textsuperscript{\rm 2},
    Xiaohui Lv\textsuperscript{\rm 1}
}
\affiliations{
    ByteDance\textsuperscript{\rm 1} \quad 
    Sun Yat-sen University\textsuperscript{\rm 2} \quad
    Independent Researcher\textsuperscript{\rm 3}  \quad
    Dalian University of Technology\textsuperscript{\rm 4} 
     \\ 
    \{xuyuanfeng.123, zehuidai,  lvxiaohui\}@bytedance.com, jiapengguan@mail.dlut.edu.cn,\\
    \{liangjianzb12, wanggrun\}@gmail.com, linliang@ieee.org
}

\begin{document}

\maketitle

\renewcommand{\thefootnote}{\fnsymbol{footnote}} 
\footnotetext[2]{Corresponding authors.} 
\renewcommand{\thefootnote}{\arabic{footnote}} 

\begin{abstract}

Small Language Models (SLMs) are a cost-effective alternative to Large Language Models (LLMs), but often struggle with complex reasoning due to their limited capacity and a tendency to produce mistakes or inconsistent answers during multi-step reasoning. Existing efforts have improved SLM performance, but typically at the cost of one or more of three key aspects: (1) \textbf{reasoning capability}, due to biased supervision that filters out negative\footnote{In this work, we refer to reasoning paths that lead to the correct answer as positives, and those that do not as negatives.} reasoning paths and limits learning from errors; (2) \textbf{autonomy}, due to over-reliance on externally generated reasoning signals; and (3) \textbf{generalization}, which suffers when models overfit to teacher-specific patterns.
In this paper, we introduce \textbf{ReaLM}, a reinforcement learning framework for robust and self-sufficient reasoning in vertical domains. To enhance reasoning capability, we propose Multi-Route Process Verification (\textbf{MRPV}), which contrasts both positive and negative reasoning paths to extract decisive patterns. To reduce reliance on external guidance and improve autonomy, we introduce Enabling Autonomy via Asymptotic Induction (\textbf{EAAI}), a training strategy that gradually fades external signals. To improve generalization, we apply guided chain-of-thought distillation to encode domain-specific rules and expert knowledge into SLM parameters, making them part of what the model has learned.
Extensive experiments on both vertical and general reasoning tasks demonstrate that ReaLM significantly improves SLM performance across aspects (1)–(3) above.

\end{abstract}

\begin{figure}[t]
  \centering
\includegraphics[width=1.0\linewidth]{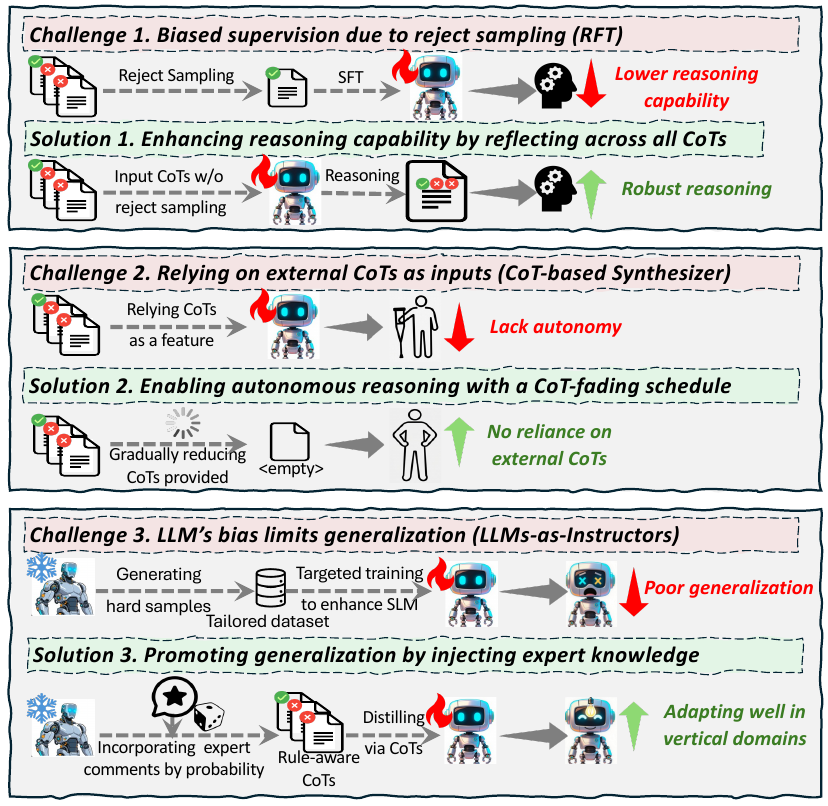}
  
  \caption{\small{\textbf{Comparison with typical SLM reasoning methods.} 
(i) RFT uses only correct CoTs, limiting learning from errors (\emph{capability}). 
(ii) CoT-based Synthesizer requires external CoTs at inference, reducing independence (\emph{autonomy}). 
(iii) LLMs-as-Instructor tailors data but struggles in new domains (\emph{generalization}). 
(iv) ReaLM addresses all three by learning from contrastive CoTs, fading external signals, and distilling expert knowledge.
\emph{For clarity, prompt templates are omitted, and only external CoTs are shown.}
  }}
  \label{fig:comparisons}
  \vspace{-2pt}
\end{figure}

\section{Introduction}
\label{sec:intro}

Large Language Models (LLMs)~\cite{touvron2023llama,jaech2024openai,guo2025deepseek} have demonstrated strong performance in complex reasoning tasks across a wide range of domains~\cite{suzgun2022challenging,dave2024investigating,li2024gsm}. However, their substantial training and inference costs limit their practicality in real-world deployments. This has motivated growing interest in empowering Small Language Models (SLMs, typically with $\leq$ 7B parameters)~\cite{kaplan2020scaling, liao2024textit} with comparable reasoning capabilities. SLMs are significantly more efficient and easier to deploy, especially in vertical-domain applications\footnote{Vertical domains are specialized areas within an industry or sector, defined by domain-specific knowledge and requirements.
}. However, due to their limited capacity, SLMs often produce errors or inconsistencies in multi-step reasoning.

While recent approaches have materially boosted SLM reasoning, the improvements rarely come for free. That is, they are typically secured by compromising at least one of three pillars vital for real-world deployment:
\begin{itemize}
    \item \textbf{Reasoning capability:} Many methods rely on biased supervision that filters out negative reasoning paths, as seen in RFT~\cite{yuan2023scalingrelationshiplearningmathematical}, which uses reject sampling to retain only high-confidence CoTs. This limits the model’s ability to learn from errors, resulting in brittle reasoning behavior and reduced robustness.
    
    \item \textbf{Autonomy:} Some approaches depend on externally generated reasoning steps (CoTs\footnote{External CoTs are intermediate reasoning steps (without final answers) from external LLM, used as input features to guide the SLM-distinguish from CoTs generated internally by the SLM itself.})~\cite{zhang2025cot} to guide the SLM during inference. These CoTs, generated by large teacher models, are used as input features, introducing strong auxiliary dependence and preventing the SLM from reasoning independently — especially in standalone or real-time applications.

    \item \textbf{Generalization:} Methods like LLMs-as-Instructor~\cite{ying2024llms}, which generate hard samples from SLM errors, often restrict learning to the reasoning patterns of teacher models. This teacher-specific bias causes poor generalization to domains where the teacher underperforms or where domain-specific rules differ from those seen during pretraining.
\end{itemize}
These key limitations — lack of robust reasoning capability, limited autonomy, and poor generalization — are summarized in Fig.~\ref{fig:comparisons}. Addressing all three is essential for building practical, self-sufficient reasoning systems using SLMs.

To address these challenges, we propose \textbf{ReaLM}, a reinforcement learning framework for robust and self-sufficient reasoning in domain-specific settings. ReaLM tackles all three issues through three complementary components:

\textbf{(1) Multi-Route Process Verification (MRPV)} improves reasoning capability by encouraging models to reflect on both positive and negative reasoning paths. Unlike prior work that aggregates CoTs indiscriminately \cite{zhang2025cot}, MRPV introduces structured supervision that contrasts multiple reasoning trajectories, helping the model identify key decision factors and avoid systematic failure modes.

\textbf{(2) Enabling Autonomy via Asymptotic Induction (EAAI)} promotes autonomy by gradually reducing reliance on external guidance during training. Inspired by curriculum learning~\cite{xu2025twt,liao2024textit}, EAAI enables the model to learn to reason independently in an end-to-end manner, without the need for external CoTs at inference time.

\textbf{(3) Guided CoT Distillation with Benchmarking} enhances generalization. We introduce an industrial benchmark, \textit{Search Ad Relevance Prediction}, which requires fine-grained reasoning under strict domain constraints. Guided CoT distillation enables the model to encode domain-specific rules and expert knowledge into its parameters, making this knowledge part of what the model has learned.

\paragraph{Contributions.} Our main contributions are:
\begin{itemize}
    \item We propose \textbf{ReaLM}, a reinforcement learning framework that improves SLM reasoning while addressing the three core limitations: capability, autonomy, and generalization.
    \item We introduce \textbf{MRPV}, a supervision mechanism that contrasts both positive and negative reasoning paths to extract decisive patterns, enabling robust and reflective reasoning.
    \item We design \textbf{EAAI}, a progressive training strategy that reduces reliance on external signals, allowing the model to reason independently.
    \item We construct a new industrial benchmark and apply guided CoT distillation to help SLMs acquire domain-specific knowledge without requiring explicit rule inputs.
\end{itemize}

\begin{figure*}[h]
  \centering
\includegraphics[width=\textwidth]{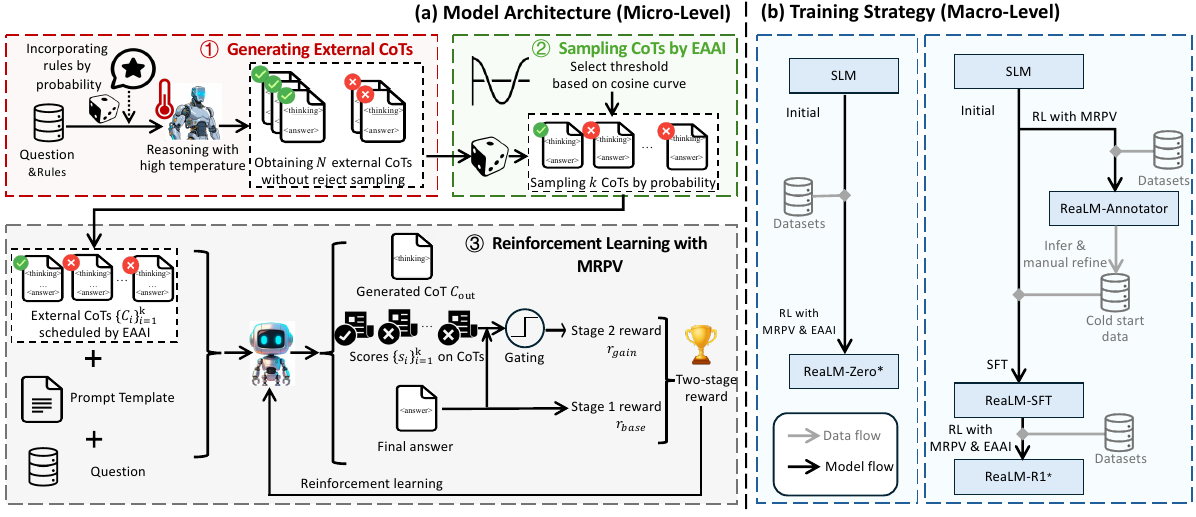}
  \caption{\textbf{Overview of ReaLM.} (a) The architecture comprises three stages. {\blacka \quad } External CoTs are generated with optional rule injection. {\blackb \quad } EAAI samples $k$ CoTs via a guidance-fading schedule. {\blackc \quad } MRPV performs RL with two key designs: (i) scoring each CoT alongside typical reasoning, and (ii) applying a two-stage reward to encourage comparison and reflection.
(b) It supports two strategies: ReaLM-Zero (pure RL) and ReaLM-R1 (iterative training).
  }
  \label{fig:overview}
\end{figure*}

\section{Method}
\label{sec:method}

Fig.~\ref{fig:overview}(a) illustrates the overall architecture. We describe ReaLM through three components aligned with our core contributions: Multi-Route Process Verification (MRPV) for robust reasoning (detailed in Sec.~\ref{sec:MRPV} and Fig.~\ref{fig:overview}(a).{\blackc \quad }), Enabling Autonomy via Asymptotic Induction (EAAI) for self-sufficiency (detailed in Sec. \ref{sec:ARAI} and Fig.~\ref{fig:overview}(a).{\blackb \quad }), and Guided CoT Distillation with an industrial benchmark for domain generalization (detailed in Sec. \ref{sec:datasets} and Fig.~\ref{fig:overview}(a).{\blacka \quad }).

Fig.~\ref{fig:overview}(b) outlines the training pipeline. 
A pure reinforcement learning strategy and an iterative training strategy are discussed separately in Sec.~\ref{sec:training}.

\subsection{Multi-Route Process Verification}
\label{sec:MRPV}

To enhance reasoning capability, we introduce MRPV, which explicitly contrasts positive and negative reasoning paths and rewards effective reflection. It comprises multi-CoT input construction and a two-stage reward mechanism.

\paragraph{Multi-CoT Input Construction.} As illustrated in 
Fig.~\ref{fig:overview}(a).{\blacka \quad }, for a given question, an LLM is queried $N$ times using temperature sampling to produce a set of reasoning paths $\{C_i\}_{i=1}^N$. These CoTs are not filtered, thus incorporating both correct and incorrect reasoning trajectories. Despite leading to wrong answers, incorrect CoTs reflect alternative perspectives that enrich the inference.

After that, a subset of $k \leq N$ CoTs is sampled in Fig.~\ref{fig:overview}(a).{\blackb \quad \ }and then included as input to the SLM in Fig.~\ref{fig:overview}(a).{\blackc \quad }. The SLM then produces its own chain-of-thought $C_{\text{out}}$ and final answer $pred$, while also assigning a utility score to each $C_i$. The outputs ($pred$ and utility scores) are used to guide model optimization through the two-stage reward function described below.

\paragraph{Two-Stage Reward Design.}
To reinforce both answer correctness and reasoning quality, we introduce a two-stage reward mechanism that encourages explicit reflection:
\begin{itemize}
    \item \textbf{Stage 1 (Answer Supervision):} Rewards the model based on correctness and format of the final answer.
    \item \textbf{Stage 2 (Process Supervision):} Assesses the utility of each external CoT by comparing its conclusion to ground truth, guiding ReaLM to reflect on valuable paths. 
\end{itemize}

Formally, the reward is defined as:

\begin{footnotesize}
\begin{align}
r &= r_{\text{base}} + r_{\text{gain}}, \label{eq:reward_1} \\
r_{\text{base}} &= 
\begin{cases}
1, & \text{if } pred \text{ and format are both correct} \\
0.1, & \text{if format is correct but answer is wrong} \\
0, & \text{otherwise}
\end{cases} \label{eq:reward_2} \\
r_{\text{gain}} &= 
\begin{cases}
scale_{\text{gain}} \cdot \frac{1}{k} \sum_{i=1}^k s_i, & \text{if } r_{\text{base}} = 1 \\
0, & \text{otherwise}
\end{cases} \label{eq:reward_3}
\end{align}
\end{footnotesize}

Here, $s_i \in \{0, 1\}$ denotes whether the SLM's assessment of $C_i$ matches the conclusion drawn within $C_i$. Stage 2 activates only if Stage 1 succeeds, ensuring training stability.

Unlike simply concatenating multiple CoTs — which tends to encourage the model to summarize shared reasoning patterns — we explicitly contrast the differences among CoTs through a two-stage reward mechanism, in order to highlight critical decision factors.

\paragraph{Comparison with GRPO.}

While both MRPV and GRPO leverage multiple reasoning trajectories, they differ fundamentally in design. GRPO treats each query-CoT pair as a separate data point and aggregates rewards across a batch of such examples. MRPV, in contrast, integrates all CoTs directly into a single input context and trains the model to reason jointly over these references. This enables token-level interactions and contrastive analysis between reasoning paths, facilitating more reflective and robust inference.

\paragraph{Theoretical Motivation.}
We formally analyze the benefit of leveraging diverse CoT references to enhance reasoning performance. Specifically, we demonstrate that incorporating external CoTs can lead to a higher expected reward than relying solely on reasoning generated without such references. The result is stated as follows:

\begin{mdframed}[style=MyFrame2]
\begin{restatable}[Reward Bound]{prop}{sec:method}
\label{prop:reward}
Let $C_{\text{SLM}}$ denote the reasoning path generated without access to external CoTs. Under mild assumptions (see Appendix C), the expected reward of the output $C_{\text{out}}$, generated with reference to a set of CoTs ${C_i}$, satisfies:
\begin{equation}
\mathbb{E}[R(C_{\text{out}})] \geq \mathbb{E}[R(C_{\text{SLM}})]. \label{eq:reward_gap}
\end{equation}
\end{restatable}
\end{mdframed}

A proof sketch of Prop. \ref{prop:reward} is provided in Appendix C. 

\vspace{10pt}
Beyond bounding the performance of $C_{\text{out}}$, we also characterize its reward by the underlying contributions of the referenced CoTs. In particular, the following result establishes that the reward of $C_{\text{out}}$ can be represented as a learnable weighted combination of the individual CoT rewards.

\vspace{11pt}
\begin{mdframed}[style=MyFrame2]
\begin{restatable}[Reward Decomposition]{prop}{sec:method}
\label{prop:decomposition}
Under the same assumptions as Proposition~\ref{prop:reward}, there exists a set of learnable weights $\{w_i\}_{i=1}^k$, where 
$\sum_{i=1}^k w_i = 1$, such that the expected reward of the output $C_{\text{out}}$ can be expressed as:
\begin{equation}
\mathbb{E}[R(C_{\text{out}})] = \mathbb{E}\left[R(\sum_{i=1}^{k} w_i \cdot C_i)\right] + \Delta_{\text{gen}}, \label{eq:reward_decomposition}
\end{equation}
where $\Delta_{\text{gen}} \geq 0$ denotes additional gains from novel or refined generation by the SLM beyond direct imitation of the references.
\end{restatable}
\end{mdframed}

This decomposition highlights the importance of contrasting all reference CoTs, including both high-quality (positive) and low-quality (negative) paths. Learning appropriate weights over $\{C_i\}$ enables the model to selectively amplify informative reasoning while suppressing misleading cues. This justifies the necessity of our MRPV. A proof sketch of Prop. \ref{prop:decomposition} is also included in Appendix C.

\subsection{Enabling Autonomy via Asymptotic Induction}
\label{sec:ARAI}

MRPV improves reasoning, but relies on external CoTs. To support standalone reasoning, we introduce EAAI as shown in Fig.~\ref{fig:overview}(a).{\blackb \quad }, a curriculum-based fading mechanism that gradually removes reliance on external CoTs during training.

\paragraph{CoT Sampling with Cosine Decay.}
During training, external CoTs are included via a probability decaying over time:

\begin{equation}
\text{threshold} = 0.5 \cdot (1 + \cos(\pi \cdot t)), \quad t = \frac{\text{step}}{\text{total\_steps}}.
\end{equation}
Only CoTs with sampled probability below this threshold are used, ensuring a smooth transition toward independent reasoning. The full algorithm is in Algorithm~\ref{alg:algorithm_cot}.

\begin{algorithm}[H]
\SetAlgoNoEnd
\begin{footnotesize}
\KwIn{$cur$: current training step\\
\hspace{29pt}$total$: total training steps\\
\hspace{29pt}$Candidates$: set of candidate CoTs (Size $N$)\\}
\KwOut{$Selected$: selected subset of CoTs}

\hspace{8pt} $t$ $\leftarrow$ $cur$ / $total$;

\hspace{8pt} $threshold$ $\leftarrow$ 0.5 $\cdot$ (1 + cos($\pi$ $\cdot$ $t$));

\hspace{8pt} $Selected \leftarrow \emptyset$;

\hspace{8pt} \For{$i \in \{1,2,...,N\}$}{
\hspace{8pt} $prob$ $\leftarrow$ Uniform(0,1);

\hspace{8pt} \If{$prob<threshold$}{
\hspace{8pt} $c$ $\leftarrow$ RandomChoice($Candidates$);

\hspace{8pt} $Selected \leftarrow Selected \cup \{ c \}$;

\hspace{8pt} $Candidates \leftarrow Candidates \setminus \{ c \}$
}

}

\hspace{8pt} \textbf{return} $Selected$;
\end{footnotesize}

\caption{Dynamic CoT selection via Cosine decay}
\label{alg:algorithm_cot}
\end{algorithm}

\paragraph{Reward Stability.}
When the sampling probability becomes too low, we disable Stage 2 rewards from MRPV to avoid reward sparsity, using only Stage 1 to guide optimization.

\subsection{Improving Generalization via Guided CoT Distillation and Benchmarking}
\label{sec:datasets}

To help the model generalize beyond teacher-specific reasoning and adapt to real-world domains, we distill domain-specific rules into the SLM using a proprietary dataset.

\paragraph{Industrial Benchmark: Ad Search Relevance Prediction.}
We introduce a high-quality vertical dataset from a commercial search system. Each data point is a tuple $\langle query, ad\_info, label, comment \rangle$, where the $comment$ encodes expert rationale and proprietary rules not in public corpora. The $label$ is a ternary relevance score (irrelevant, partially relevant, relevant). Following complex proprietary rules, it represents distributions beyond standard LLM pretraining, serving as a challenging real-world benchmark. See Appendix D for available examples.

\paragraph{Rule-Aware CoT Distillation.}
To help the model encode domain-specific rules and knowledge in its parameters, in Fig.~\ref{fig:overview}(a).{\blacka \quad }, we generate rule-aware CoTs using a general teacher LLM by injecting expert comments into prompts with a certain probability. These rule-enhanced CoTs are then sampled for the following training.

\subsection{Training Paradigm}
\label{sec:training}

As shown in Fig.~\ref{fig:overview}(b), we train two variants following the strategy of \cite{guo2025deepseek}:

\begin{itemize}
    \item \textbf{ReaLM-Zero} (left of Fig.~\ref{fig:overview}(b)): Trained with reinforcement learning using MRPV and EAAI. It serves as a lightweight, fully autonomous baseline.
    \item \textbf{ReaLM-R1} (right of Fig.~\ref{fig:overview}(b)): Built through iterative training: ReaLM-Annotator first labels cold-start data for SFT, followed by RL to progressively enhance reasoning with stable initialization.

\end{itemize}

\section{Experiments}
\label{sec_ex}

\subsection{Experiment Setup} 

\paragraph{Datasets.} Following \cite{ying2024llms}, we evaluated our methodology on mathematical reasoning, factual reasoning, and general complex reasoning tasks. Furthermore, to assess reasoning capability on vertical domains that require specific knowledge, we tested the model on academic and industrial semantic matching tasks requiring reasoning ability.

\begin{itemize}
 \item \textbf{Mathematical reasoning}: We chose  Grade School Math 8K (GSK8K) \cite{cobbe2021training} dataset as an in-domain (ID) dataset. We also tested the model on out-of-domain (OOD) GSK8KPlus \cite{li2024gsm} dataset.

 \item \textbf{General complex reasoning}: We selected  BIGBench Hard (BBH) \cite{suzgun2022challenging} as an ID dataset. The OOD datasets include AGIEval \cite{zhong2023agieval} and BIGBench \cite{srivastava2023beyond} filtered by CasCoD (BB-sub) \cite{dai2024improve}.

  \item \textbf{Factual reasoning}: We used Multitask Language Understanding (MMLU) \cite{hendrycks2020measuring} as an ID dataset. The OOD datasets include AI2 Reasoning Challenge (ARC-C) and Easy (ARC-E) \cite{clark2018think}.

 \item \textbf{Semantic matching}: We employed Microsoft Research Paraphrase Corpus (MRPC) \cite{dolan2005automatically} as the dataset. We also tested the model on a private industrial semantic and matching dataset, i.e., Ad Search Relevance Prediction, requiring reasoning ability on specific rules.
\end{itemize}

The details of these tasks are described in Appendix A. For academic benchmark datasets, we report the accurate match \cite{rajpurkar2016squad} for comparison. For the proprietary industrial dataset, Ad Search Relevance Prediction, we evaluate both accuracy and F1 score on the non-zero classes, aligning with practical usage scenarios in commercial systems.

\begin{table*}[t]
  \centering
  \resizebox{\linewidth}{!}{
      \begin{tabular}{cccccccccccc}
        \toprule
         \multirow{2}{*}{\makecell{\textbf{External CoTs}\\ \textbf{Provided}}}& \multirow{2.5}{*}{\textbf{Methods}} & \multicolumn{5}{c}{\textbf{In-Domain}}& \multicolumn{4}{c}{\textbf{Out-Of-Domain}} & \multirow{2}{*}{\textbf{Avg}} \\
         
        \cmidrule(r){3-6}\cmidrule(r){7-11}
         & & \textbf{GSM8K} & \textbf{BBH} & \textbf{MMLU} & \textbf{MRPC} & \textbf{GSM8KPlus} & \textbf{AGIEval}  & \textbf{BB-sub} & \textbf{ARC-C} & \textbf{ARC-E} & \\
        \midrule
        
        \rowcolor{gray!50} \multicolumn{12}{l}{\textit{\# Deepseek-R1-7B}}\\

         & usc  &  93.1  &  85.0  &  91.1  &  86.1  &  82.7  &  54.9  &  46.8  &  79.9  &  88.2  &  78.7  \\
        w/ & CoT-synthesizer  & 93.8  &  87.2  &  92.1  &  87.3  &  87.0  &  58.7  &  46.3  &  83.6  &  90.6  &  80.7  \\
        & ReaLM w/o EAAI & \textbf{96.0}\scalebox{0.6}{\(\pm\)0.3}  
        &\textbf{92.2}\scalebox{0.6}{\(\pm\)0.4}
        &\textbf{95.0}\scalebox{0.6}{\(\pm\)0.1}
        &\textbf{91.8}\scalebox{0.6}{\(\pm\)0.4}
        &\textbf{92.5}\scalebox{0.6}{\(\pm\)0.7}
        &\textbf{61.8}\scalebox{0.6}{\(\pm\)0.9}
        &\textbf{49.0}\scalebox{0.6}{\(\pm\)0.4}
        &\textbf{90.3}\scalebox{0.6}{\(\pm\)0.4}
        &\textbf{95.5}\scalebox{0.6}{\(\pm\)0.5}
        &\textbf{84.9}\scalebox{0.6}{\(\pm\)0.5}
        \\

        \hdashline

        \rowcolor{gray!20}  & SFT  &  30.1  &  70.3  &  56.8  &  78.9  &  22.8  &  38.7  &  40.3  &  73.4  &  83.9  &  55.0  \\
        \rowcolor{gray!20}  & MT-CoT  &  81.0  &  73.9  &  63.6  &  85.3  &  73.5  &  36.6  &  43.5  &  73.4  &  82.3  &  68.1  \\
        \rowcolor{gray!20}  w/o& SKIntern  &  85.5  &  74.5  &  66.5  &  85.5  &  78.5  &  42.0  &  39.2  &  76.9  &  85.7  &  70.5  \\
        \rowcolor{gray!20}  & RL  & 92.3  &  85.1  &  74.3  &  85.0  &  \textbf{91.2}  &  55.0  &  45.6  &  83.8  &  91.8  &  78.2  \\
        \rowcolor{gray!20}  & ReaLM  &  \textbf{94.4}\scalebox{0.6}{\(\pm\)0.5}  
        &\textbf{88.9}\scalebox{0.6}{\(\pm\)0.4}  
        &\textbf{77.3}\scalebox{0.6}{\(\pm\)0.3}
        &\textbf{88.0}\scalebox{0.6}{\(\pm\)0.2}
        &90.8\scalebox{0.6}{\(\pm\)0.5}
        &\textbf{58.4}\scalebox{0.6}{\(\pm\)0.8}
        &\textbf{48.7}\scalebox{0.6}{\(\pm\)0.7}
        &\textbf{87.7}\scalebox{0.6}{\(\pm\)0.6}
        &\textbf{93.1}\scalebox{0.6}{\(\pm\)0.7}
        &\textbf{80.8}\scalebox{0.6}{\(\pm\)0.6}
        \\
        
        \midrule
        \rowcolor{gray!50} \multicolumn{12}{l}{\textit{\# Qwen2.5-7B}}\\

         &  usc  & 87.0  &  82.3  &  90.1  &  87.7  &  82.6  &  65.2  &  45.4  &  87.4  &  89.6  &  79.7  \\
         w/ & CoT-synthesizer  &  93.6  &  87.3  &  92.8  &  87.6  &  85.2  &  69.1  &  48.6  &  89.1  &  95.3  &  83.2  \\
         & ReaLM w/o EAAI 
         &\textbf{95.1}\scalebox{0.6}{\(\pm\)0.2} &\textbf{91.4}\scalebox{0.6}{\(\pm\)0.3}
         &\textbf{95.3}\scalebox{0.6}{\(\pm\)0.1}
         &\textbf{92.6}\scalebox{0.6}{\(\pm\)0.1}
         &\textbf{90.6}\scalebox{0.6}{\(\pm\)0.4}
         &\textbf{72.0}\scalebox{0.6}{\(\pm\)1.2}
         &\textbf{53.2}\scalebox{0.6}{\(\pm\)0.7}
         &\textbf{96.7}\scalebox{0.6}{\(\pm\)0.3}
         &\textbf{98.6}\scalebox{0.6}{\(\pm\)0.4}
         &\textbf{87.3}\scalebox{0.6}{\(\pm\)0.5}  \\

        \hdashline
        
        \rowcolor{gray!20} & SFT  & 28.8  &  73.4  &  70.5  &  84.1  &  23.6  &  44.0  &  41.3  &  84.7  &  87.2  &  59.7  \\
        \rowcolor{gray!20} & MT-CoT  &  79.3  &  76.6  &  72.7  &  85.4  &  73.4  &  40.6  &  39.8  &  82.5  &  86.2  &  70.7  \\
        \rowcolor{gray!20} w/o & SKIntern  &  85.5  &  76.9  &  74.0  &  86.2  &  76.2  &  45.9  &  42.3  &  87.0  &  89.1  &  73.7  \\
        \rowcolor{gray!20} & RL  & 91.5  &  85.8  &  73.2  &  86.3  &  88.6  &  57.9  &  45.3  &  88.1  &  95.1  &  79.1  \\ 
        \rowcolor{gray!20} & ReaLM &\textbf{92.3}\scalebox{0.6}{\(\pm\)0.4}
        &\textbf{88.5}\scalebox{0.6}{\(\pm\)0.3}  &\textbf{77.7}\scalebox{0.6}{\(\pm\)0.3}
        &\textbf{88.9}\scalebox{0.6}{\(\pm\)0.2}
        &\textbf{89.6}\scalebox{0.6}{\(\pm\)0.4}
        &\textbf{60.9}\scalebox{0.6}{\(\pm\)0.6}
        &\textbf{51.4}\scalebox{0.6}{\(\pm\)0.6}
        &\textbf{89.3}\scalebox{0.6}{\(\pm\)0.5}
        &\textbf{95.9}\scalebox{0.6}{\(\pm\)0.4}
        &\textbf{81.6}\scalebox{0.6}{\(\pm\)0.4}  \\
        
        \bottomrule
      \end{tabular}
      }
    \caption{\textbf{Comparison between ReaLM and other advanced methods.} Initialized with Deepseek-R1-7B and Qwen2.5-7B, ReaLM consistently demonstrates superior performance. The best scores under each setting are highlighted in bold. ``External CoTs Provided" indicates whether external CoTs are available during inference (aligned with Tab.~\ref{table5}). For ReaLM, results are reported as the mean and standard deviation of accuracy over three independent runs.
    }
    \label{table_1}
\end{table*}

\begin{table}[t]
\centering
\resizebox{1.0\linewidth}{!}{
    \begin{tabular}{ccccc}
    \toprule
    \multirow{2.7}{*}{\makecell{ Variants}} & \multicolumn{3}{c}{Training} & Inference \\ 
     \cmidrule(r){2-4}\cmidrule(r){5-5}
     & \makecell{External \\CoT} & \makecell{Two-Stage \\Reward} & \makecell{CoT \\ Decay} & \makecell{ External \\ CoT} \\ 
    \midrule
    MR & $\checkmark$ &  &  & $\checkmark$ \\ 
    MRPV & $\checkmark$ & $\checkmark$ &  & $\checkmark$ \\ 
    MR+EAAI & $\checkmark$ &  & $\checkmark$ &   \\ 
    \makecell{ReaLM: MRPV+EAAI} & $\checkmark$ & $\checkmark$ & $\checkmark$ &   \\ 
    \bottomrule
    \end{tabular}
}
\caption{\textbf{Components used during training} for four variants. ARAI enables external-CoT-free reasoning at inference.
}
\label{table7}
\end{table}

\begin{table}[t]
  \centering
  \resizebox{1.0\linewidth}{!}{
  \begin{tabular}{cccccc}
    \toprule
    \multirow{2}{*}{\makecell{External  CoTs \\ Provided}} & \multirow{2}{*}{Varients}&
        \multicolumn{2}{c}{\makecell{Deepseek-R1-7B}} 
        &\multicolumn{2}{c}{Qwen2.5-7B} \\
        \cmidrule(r){3-4}\cmidrule(r){5-6}
    & & ID & OOD & ID & OOD \\
    \midrule
    \multirow{2}{*}{w/} &  MR  &  93.1  &  76.1  &  92.9  &  81.3 \\
     &  MRPV  &  93.7  &  77.8  &  93.6  &  82.2 \\
    \hdashline
    \rowcolor{gray!20} & RL (baseline) & 84.2& 73.5& 84.2& 75.0\\
    \rowcolor{gray!20} wo  & MR + EAAI  &  86.0  &  74.0  &  85.8  &  76.2 \\
     \rowcolor{gray!20} 
     & MRPV + EAAI  &  87.1  &  75.8  &  86.9  &  77.4 \\
    \bottomrule
    \end{tabular}
  }
  \caption{\textbf{Ablation studies} on GSM8K, initialized with Deepseek-R1-7B and Qwen2.5-7B.
  }
  \label{table5}
\end{table}

\begin{table}[t]
\setlength{\abovecaptionskip}{2mm}
  \centering
  \resizebox{\linewidth}{!}{
    \begin{tabular}{cccc}
    \toprule
    Variants  &Total Err$\downarrow$ & \makecell{SAME  Err$\downarrow$} & \makecell{DIFF  Err $\downarrow$} \\
    \midrule
    \rowcolor{gray!20} \multicolumn{4}{l}{\textit{\# RL Approach}}\\
    RL    &7.7 &5.5 &2.2 \\
    \midrule
    \rowcolor{gray!20} \multicolumn{4}{l}{\textit{\# External CoTs Provided (without EAAI)}}\\
     MR     &4.6 &2.3 &2.3 \\
     MRPV    &\textbf{4.0} &\textbf{1.8} &2.2 \\
     MRPV-Reject    &5.1 &2.8&2.3 \\
    \midrule
    \rowcolor{gray!20} \multicolumn{4}{l}{\textit{\# No External CoT is Provided (with EAAI)}}\\
      MR    &6.7 &4.7 &2.0 \\
      MRPV    &\textbf{5.6} &\textbf{3.7} &2.0 \\
      MRPV-Reject    &7.1 &5.2 &2.0 \\
    \bottomrule
    \end{tabular}
    }
  \caption{\textbf{Error analysis of SLM predictions} on the GSM8K dataset, categorized by whether they match any of the LLM predictions. Variants are initialized with Deepseek-R1-7B. ``MRPV-Reject" denotes variants that apply reject sampling external CoTs before training.
  }
  \label{table2}
\end{table}

\paragraph{Baselines.} To rigorously evaluate ReaLM on the above reasoning tasks, we benchmark it against a diverse suite of SOTA baselines spanning multiple domains, including methods that operate both with/without external CoTs, and trained under SFT as well as RL paradigms, as detailed below:

\begin{itemize}
    \item Supervised-Finetuning (SFT) on the final ground truth.
    
    \item Multitask-CoT-Distillation (MT-CoT)~\cite{li2022explanations}, using multi-task learning to distill reasoning ability.

    \item SKIntern~\cite{liao2024textit}, empowering SLMs to internalize symbolic knowledge gradually.

   \item Universal Self-Consistency(USC)~\cite{wang2022self,chen2023universal}, sampling a diverse set of reasoning paths, and then selects the most consistent answer.

   \item CoT-Synthesizer~\cite{zhang2025cot}, leveraging CoT reasoning to synthesize answers by analyzing complementary information from multiple candidate responses.

   \item RL with Group Relative Policy Optimization (GRPO) \cite{shao2024deepseekmath} to equip reasoning capability.
   
\end{itemize}

\paragraph{Implementation Details.}
We queried GPT-4-turbo with a temperature of 1.5 to generate diverse CoTs and predictions. Examples are in Appendix D. For fair comparison, student SLMs are initialized with two pretrained models, DeepSeek-R1-7B \cite{guo2025deepseek} and Qwen2.5-7B~\cite{hui2024qwen2}, to reduce bias from single initialization. All methods share identical experimental settings, including external CoTs and initialization weights when applicable.

For supervised-finetuning (SFT) approaches, we use full-parameter tuning with DeepSpeed ZeRO Stage 2 \cite{rajbhandari2020zero}. For RL approaches, we employ GRPO with rule-based reward for CoT judgements and predictions.  All experiments were conducted on 8 A100 GPUs with 80GB. More detailed information about training, inference, and hyperparameters are in  Appendix B.

\subsection{Main Results}
EAAI enables ReaLM to perform inference without relying on external CoTs. Without EAAI, ReaLM still requires external CoT inputs. Tab.~\ref{table_1} reports the performance of advanced models and ReaLM, grouped by whether external CoTs are provided during inference.

Under the setting where external CoTs are used (white rows), ReaLM without EAAI outperforms CoT-Synthesizer and other CoT-input-based methods, achieving accuracy gains of 4.2\% and 4.1\% on Deepseek-R1-7B and Qwen2.5-7B, respectively. These results set a new state-of-the-art among CoT-input-based approaches, demonstrating the superior synthesis capability of MRPV.

Under the setting where no external CoTs are provided (gray rows), the full ReaLM still achieves strong performance, with average accuracy improvements of 2.6\% and 2.5\% on Deepseek-R1-7B and Qwen2.5-7B, respectively, further validating the effectiveness of our proposed method.

\subsection{Analysis}

\paragraph{Ablation Studies.}

To investigate the impact of process verification, that is,
the additional reward $r_{gain}$ for external CoT judgment in Eq.~\ref{eq:reward_3}, we perform
ablation studies by removing process verification.

Tab.~\ref{table7} presents four variants in the ablation study, each applying different subsets of our proposed components during training. The full ReaLM, includes both MRPV and EAAI. Removing process verification from MRPV while retaining multi-CoTs during training yields the MR variant. Notably, variants equipped with the EAAI module during training are capable of performing inference without external CoT.

In Tab.~\ref{table5}, the ablation results are grouped based on whether the EAAI module is used, that is, whether external CoTs are required during inference. Within each group, we examine the performance impact of removing process verification. It shows that process verification consistently improves performance by approximately 1\% on both ID and OOD datasets, confirming its effectiveness and that enabling ReaLM to assess external CoTs' utility boosts reasoning accuracy.

In addition, the variant of MR+EAAI in Tab.~\ref{table5} significantly outperforms the RL method, indicating that, beyond process verification, exposing the model to multiple reasoning paths during training further enhances SLM's performance. This supports Proposition~\ref{prop:reward} and further confirms that external CoTs contribute to performance gains.

Moreover, in Tab.~\ref{table_1}, ReaLM (gray row) shows a performance drop compared to ReaLM without EAAI (white row). This suggests that while EAAI enables adaptation to inference without external CoTs and retains part of the performance gain, there is still room for improvement.

\paragraph{Learning from Failure Rational.}

\begin{figure}[h]
  \centering
  \includegraphics[width=1.0\linewidth]{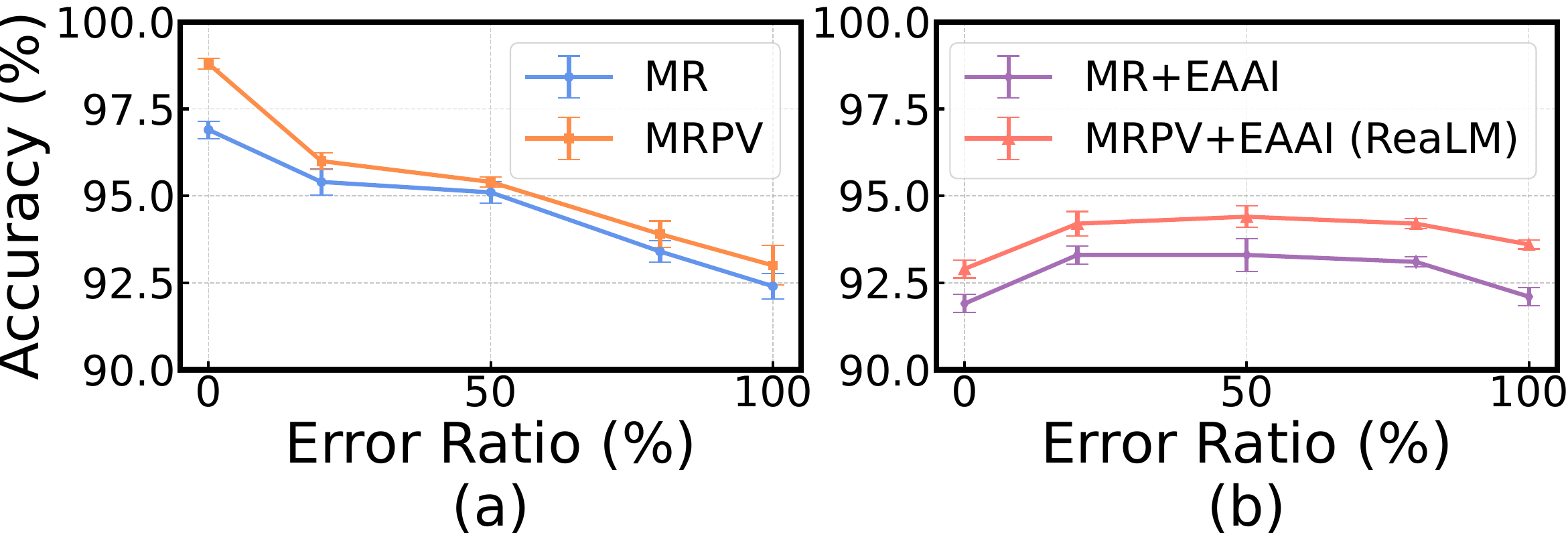}
  \caption{\textbf{Impact of Incorrect CoT Ratio on Performance.} Variants are initialled with Deepseek-R1-7B and evaluated on GSM8K.
  (a) With external CoTs provided during inference, and (b) without external CoTs provided during inference.
  }
  \label{fig:ablationOnErrorRate}
\end{figure}

Fig.~\ref{fig:ablationOnErrorRate} investigates the relationship between the ratio of incorrect CoTs provided to ReaLM and the performance of ReaLM.

The error ratio of external CoTs is defined as the proportion of incorrect CoTs among all provided ones, where correctness is determined by whether the CoT leads to a prediction consistent with the ground truth.

In Fig.~\ref{fig:ablationOnErrorRate}(a), with reject sampling (0\% error ratio), ReaLM without EAAI achieves a notable performance gain. However, since all input CoTs are correct, this improvement comes from the use of reject sampling, rather than the model's inherent reasoning ability. Fig.~\ref{fig:ablationOnErrorRate}(a) provides a clear explanation, where the performance suffers from a steady decline as the error ratio increases in the absence of the EAAI module.

In contrast, equipped with EAAI, ReaLM in Fig.~\ref{fig:ablationOnErrorRate}(b) learns from external CoTs to enhance its reasoning ability and gradually achieves independent inference. Notably, the performance is higher when the error ratio falls within the 20\%-80\% range, compared to the extreme case of 0\% (all correct) or 100\% (all incorrect), indicating that both correct and incorrect reasoning paths contribute to training. This demonstrates the benefit of exposing the model to diverse, even imperfect, reasoning paths.

\begin{table}[h]
  \centering
  \resizebox{0.9\linewidth}{!}{
  \begin{tabular}{ccccc}
    \toprule
    Methods & Acc$\uparrow$ & F1 Score$\uparrow$ \\
    \midrule
    \rowcolor{gray!20} \multicolumn{3}{l}{\textit{\# Other Advanced Methods}}\\
    SKIntern  &  80.19  &  79.05 \\
    CoT-synthesizer &  81.54  &  80.26 \\
    \hdashline    
    \rowcolor{gray!20} \multicolumn{3}{l}{\textit{\# Different Variants of ReaLM}}\\
    ReaLM-Zero w/o Expert Comments  &  82.30  &  81.40 \\
    ReaLM-Zero  &  82.56  &  81.80\\
    ReaLM-R1  &  \textbf{82.61}  &  \textbf{82.45}\\
    \bottomrule
    \end{tabular}
  }
  \caption{\textbf{Results on our industrial benchmark.} ReaLM-R1 achieves the best performance, with gains from expert comments and iterative training.}
  \label{table6}
\end{table}

We further investigate the root cause of this phenomenon by comparing the prediction errors made by the SLM with those of the LLM. As shown in Tab.~\ref{table2}, we categorize incorrect predictions from SLM into two types: those that match any incorrect prediction from the LLM (SAME Err) and those that differ from all LLM outputs (DIFF Err). A high SAME Err reflects that error reasoning patterns are shared between the SLM and LLM. While this is not a perfect indicator of reasoning similarity, we adopt it as a practical proxy under the assumption that semantically or structurally similar reasoning paths are more likely to lead to identical mistakes.

As shown in Tab.~\ref{table2}, MRPV-Reject variants exhibit a higher SAME Err ratio, likely due to the rejection sampling mechanism discarding incorrect training examples, thereby limiting opportunities to learn from failure cases. In contrast, MRPV achieves a significantly lower SAME Err rate than MR, indicating its ability to learn from incorrect CoTs to avoid repeating similar mistakes.

Moreover, DIFF Err remains nearly unchanged across models with and without EAAI, which aligns with our analysis: the performance gains of MRPV primarily come from reducing SAME Err through effective learning from LLM failures.

\paragraph{Two-stage Reward Mechanism}

To assess the effectiveness of the two-stage reward mechanism, we remove the gating mechanism in Eq.~\ref{eq:reward_3}, such that the auxiliary reward $r_{gain}$ is applied regardless of whether the final prediction is correct. This modification leads to a performance drop of ReaLM from 94.4 to 93.8. We hypothesize that in cases where the model makes incorrect predictions, it may optimize the overall reward by focusing on $r_{gain}$ — the reward for the auxiliary task — instead of improving main task performance. In contrast, with the two-stage reward mechanism, reasoning is only enhanced when the main task prediction is correct, thus avoiding misleading updates from erroneous samples.

\paragraph{Performance on Industry Dataset}

To assess generalization to unseen domains with proprietary rules, we benchmark ReaLM variants and advanced methods on our industrial dataset. As shown in Tab.~\ref{table6}, ReaLM-Zero consistently outperforms SKIntern and CoT-Synthesizer, demonstrating the strength of our RL-based framework in handling domain-specific reasoning.

We further examine the impact of incorporating proprietary rules. Introducing expert comments boosts performance notably, indicating their utility in facilitating rule acquisition.

Finally, our iterative training pipeline produces ReaLM-R1, which surpasses ReaLM-Zero and achieves the highest overall accuracy, highlighting the benefits of cold-start distillation followed by reinforcement learning.

\section{Justifications}
\label{discussion}

This paper investigates the potential of comparative reasoning with SLMs. While prior studies have explored how to enhance the reasoning capabilities of small language models, they often involve trade-offs among reasoning capacity, autonomy, and generalization. We propose a unified framework that systematically improves all three aspects.

\textbf{Reasoning Capacity.} We show that prompting SLMs to contrast correct and incorrect reasoning paths leads to more accurate and robust inferences. While earlier work (e.g., CoT-Synthesizer) observed that flawed CoTs can occasionally yield correct answers, we go further by offering both theoretical and empirical insights. Sec.~\ref{sec:MRPV} demonstrates that exposing models to diverse CoTs and training them to assess utility encourages deeper reasoning. Tab.~\ref{table2} confirms that performance gains come from correcting flawed reasoning patterns, rather than memorizing the answer.

\textbf{Autonomy.} To reduce reliance on external supervision, we propose EAAI — a module that performs a CoT-level decay. Inspired by curriculum learning used in knowledge distillation, EAAI gradually reduces reliance on external signals during training, promoting independent reasoning.

\textbf{Generalization.} We inject expert knowledge during the LLM querying stage in Fig.~\ref{fig:overview}(a).{\blacka \quad } to incorporate domain-specific knowledge, helping SLMs generalize to unseen tasks.

While ReaLM demonstrates strong performance, there remains room for further refinement. To encourage diverse reasoning trajectories, we adopt random sampling for external CoTs. Although this approach may introduce redundancy, it avoids the domain-specific biases raised from manually crafted selection heuristics — especially in vertical domains where such biases may hinder reasoning ability. Finally, we choose the random strategy to preserve reasoning flexibility, accepting a degree of redundancy as a trade-off.

\section{Related Works}
\label{sec:rela}

\paragraph{Reasoning with SLMs.}
SLMs offer a cost-effective and flexible alternative to LLMs, prompting efforts to transfer reasoning abilities from LLMs to SLMs. For example, \cite{ying2024llms} generates challenging examples that expose SLM weaknesses, enabling targeted training. Other methods~\cite{zhou2021rethinking} further align SLMs with LLMs by supervising them with LLM-generated labels at both output and representation levels. However, these methods often depend heavily on teacher models' performance on specific tasks, limiting their generalization~\cite{ham2024difficulty}.

To overcome this, approaches~\cite{zhang2025cot} treat CoTs from LLM teachers as features, providing them as inputs to SLMs. Fine-tuning on these CoTs allows SLMs to extend the reasoning process themselves, offering greater flexibility and improved generalization to unseen tasks without explicit teacher intervention.

\paragraph{CoT Distillation.}
CoT significantly enhances LLM reasoning by breaking down complex problems into intermediate steps~\cite{wei2022chain, kim2023cot}. Recent studies~\cite{chenglin2024mixed} show that CoTs can be used to distill teacher models' reasoning abilities to smaller models, typically by using LLM-generated CoTs to fine-tune student models~\cite{chen2023mcc,liu2023mind}. Compared to traditional distillation methods that use soft labels, CoT distillation offers better performance and greater interpretability~\cite{han2023impact}.

However, SLMs' limited capacity often leads to suboptimal performance~\cite{zhang2023towards} when directly applying this method to SLMs. To address this, some studies refine LLM-generated CoTs to improve SLMs' performance on mathematical and logical reasoning tasks. For example, \cite{yuan2023scalingrelationshiplearningmathematical} applies rejection sampling to select high-confidence CoTs, aligning SLM and LLM outputs. Other methods, such as \cite{lightman2023let}, emphasize the correctness of intermediate steps to improve final performance. However, these methods primarily focus on learning from successful reasoning paths and often overlook failure cases.

In contrast, our method leverages both successful and failed reasoning trajectories, enabling the model to compare and learn from both. This broader exploration improves SLMs' reasoning robustness and generalization.

\section{Conclusion}
\label{conclusion}

In this paper, we propose ReaLM, an RL framework that enables SLMs with autonomous, robust reasoning capabilities. By learning from both successful and failed reasoning paths, fading external signal, and distilling domain knowledge, ReaLM improves reasoning ability, autonomy, and generalization -- addressing the trade-offs that constrain prior approaches. 
The effectiveness of its two core modules, MRPV and EAAI, is substantiated through theoretical analysis and targeted ablation studies.
Across a wide spectrum of vertical and general reasoning tasks, ReaLM delivers consistent gains, improving performance by 4.2\% with external CoTs and 2.6\% without them.

\bibliography{draft2026}

\begin{thebibliography}{46}
\providecommand{\natexlab}[1]{#1}

\bibitem[{Chen et~al.(2023{\natexlab{a}})Chen, Wu, Quan, Wang, Yan, and Zhang}]{chen2023mcc}
Chen, H.; Wu, S.; Quan, X.; Wang, R.; Yan, M.; and Zhang, J. 2023{\natexlab{a}}.
\newblock Mcc-kd: Multi-cot consistent knowledge distillation.
\newblock \emph{arXiv preprint arXiv:2310.14747}.

\bibitem[{Chen et~al.(2025)Chen, Chen, Li, Jiang, Wan, He, Ran, Gu, Li, Xie et~al.}]{chen2025recent}
Chen, S.; Chen, Y.; Li, Z.; Jiang, Y.; Wan, Z.; He, Y.; Ran, D.; Gu, T.; Li, H.; Xie, T.; et~al. 2025.
\newblock Recent advances in large langauge model benchmarks against data contamination: From static to dynamic evaluation.
\newblock \emph{arXiv preprint arXiv:2502.17521}.

\bibitem[{Chen et~al.(2023{\natexlab{b}})Chen, Aksitov, Alon, Ren, Xiao, Yin, Prakash, Sutton, Wang, and Zhou}]{chen2023universal}
Chen, X.; Aksitov, R.; Alon, U.; Ren, J.; Xiao, K.; Yin, P.; Prakash, S.; Sutton, C.; Wang, X.; and Zhou, D. 2023{\natexlab{b}}.
\newblock Universal self-consistency for large language model generation.
\newblock \emph{arXiv preprint arXiv:2311.17311}.

\bibitem[{Chenglin et~al.(2024)Chenglin, Chen, Li, Wang, Tao, Li, Chen, and Zhang}]{chenglin2024mixed}
Chenglin, L.; Chen, Q.; Li, L.; Wang, C.; Tao, F.; Li, Y.; Chen, Z.; and Zhang, Y. 2024.
\newblock Mixed distillation helps smaller language models reason better.
\newblock In \emph{Findings of the Association for Computational Linguistics: EMNLP 2024}, 1673--1690.

\bibitem[{Clark et~al.(2018)Clark, Cowhey, Etzioni, Khot, Sabharwal, Schoenick, and Tafjord}]{clark2018think}
Clark, P.; Cowhey, I.; Etzioni, O.; Khot, T.; Sabharwal, A.; Schoenick, C.; and Tafjord, O. 2018.
\newblock Think you have solved question answering? try arc, the ai2 reasoning challenge.
\newblock \emph{arXiv preprint arXiv:1803.05457}.

\bibitem[{Cobbe et~al.(2021)Cobbe, Kosaraju, Bavarian, Chen, Jun, Kaiser, Plappert, Tworek, Hilton, Nakano et~al.}]{cobbe2021training}
Cobbe, K.; Kosaraju, V.; Bavarian, M.; Chen, M.; Jun, H.; Kaiser, L.; Plappert, M.; Tworek, J.; Hilton, J.; Nakano, R.; et~al. 2021.
\newblock Training verifiers to solve math word problems.
\newblock \emph{arXiv preprint arXiv:2110.14168}.

\bibitem[{Dai et~al.(2024)Dai, Li, Zhou, and Hu}]{dai2024improve}
Dai, C.; Li, K.; Zhou, W.; and Hu, S. 2024.
\newblock Improve Student's Reasoning Generalizability through Cascading Decomposed CoTs Distillation.
\newblock \emph{arXiv preprint arXiv:2405.19842}.

\bibitem[{Dave et~al.(2024)Dave, Kifer, Giles, and Mali}]{dave2024investigating}
Dave, N.; Kifer, D.; Giles, C.~L.; and Mali, A. 2024.
\newblock Investigating symbolic capabilities of large language models.
\newblock \emph{arXiv preprint arXiv:2405.13209}.

\bibitem[{Dolan and Brockett(2005)}]{dolan2005automatically}
Dolan, B.; and Brockett, C. 2005.
\newblock Automatically constructing a corpus of sentential paraphrases.
\newblock In \emph{Third international workshop on paraphrasing (IWP2005)}.

\bibitem[{Geva et~al.(2021)Geva, Khashabi, Segal, Khot, Roth, and Berant}]{geva2021did}
Geva, M.; Khashabi, D.; Segal, E.; Khot, T.; Roth, D.; and Berant, J. 2021.
\newblock Did aristotle use a laptop? a question answering benchmark with implicit reasoning strategies.
\newblock \emph{Transactions of the Association for Computational Linguistics}, 9: 346--361.

\bibitem[{Guo et~al.(2025)Guo, Yang, Zhang, Song, Zhang, Xu, Zhu, Ma, Wang, Bi et~al.}]{guo2025deepseek}
Guo, D.; Yang, D.; Zhang, H.; Song, J.; Zhang, R.; Xu, R.; Zhu, Q.; Ma, S.; Wang, P.; Bi, X.; et~al. 2025.
\newblock Deepseek-r1: Incentivizing reasoning capability in llms via reinforcement learning.
\newblock \emph{arXiv preprint arXiv:2501.12948}.

\bibitem[{Ham et~al.(2024)Ham, Cho, Lee, Kang, Choi, and Kim}]{ham2024difficulty}
Ham, G.; Cho, Y.; Lee, J.-H.; Kang, M.; Choi, G.; and Kim, D. 2024.
\newblock Difficulty level-based knowledge distillation.
\newblock \emph{Neurocomputing}, 606: 128375.

\bibitem[{Han et~al.(2023)Han, Kim, Choi, and Yoon}]{han2023impact}
Han, H.; Kim, S.; Choi, H.-S.; and Yoon, S. 2023.
\newblock On the impact of knowledge distillation for model interpretability.
\newblock \emph{arXiv preprint arXiv:2305.15734}.

\bibitem[{Hendrycks et~al.(2020)Hendrycks, Burns, Basart, Zou, Mazeika, Song, and Steinhardt}]{hendrycks2020measuring}
Hendrycks, D.; Burns, C.; Basart, S.; Zou, A.; Mazeika, M.; Song, D.; and Steinhardt, J. 2020.
\newblock Measuring massive multitask language understanding.
\newblock \emph{arXiv preprint arXiv:2009.03300}.

\bibitem[{Hsieh et~al.(2023)Hsieh, Li, Yeh, Nakhost, Fujii, Ratner, Krishna, Lee, and Pfister}]{hsieh2023distilling}
Hsieh, C.-Y.; Li, C.-L.; Yeh, C.-K.; Nakhost, H.; Fujii, Y.; Ratner, A.; Krishna, R.; Lee, C.-Y.; and Pfister, T. 2023.
\newblock Distilling step-by-step! outperforming larger language models with less training data and smaller model sizes.
\newblock \emph{arXiv preprint arXiv:2305.02301}.

\bibitem[{Hui et~al.(2024)Hui, Yang, Cui, Yang, Liu, Zhang, Liu, Zhang, Yu, Lu et~al.}]{hui2024qwen2}
Hui, B.; Yang, J.; Cui, Z.; Yang, J.; Liu, D.; Zhang, L.; Liu, T.; Zhang, J.; Yu, B.; Lu, K.; et~al. 2024.
\newblock Qwen2. 5-coder technical report.
\newblock \emph{arXiv preprint arXiv:2409.12186}.

\bibitem[{Ibrahim et~al.(2024)Ibrahim, Th{\'e}rien, Gupta, Richter, Anthony, Lesort, Belilovsky, and Rish}]{ibrahim2024simple}
Ibrahim, A.; Th{\'e}rien, B.; Gupta, K.; Richter, M.~L.; Anthony, Q.; Lesort, T.; Belilovsky, E.; and Rish, I. 2024.
\newblock Simple and scalable strategies to continually pre-train large language models.
\newblock \emph{arXiv preprint arXiv:2403.08763}.

\bibitem[{Jaech et~al.(2024)Jaech, Kalai, Lerer, Richardson, El-Kishky, Low, Helyar, Madry, Beutel, Carney et~al.}]{jaech2024openai}
Jaech, A.; Kalai, A.; Lerer, A.; Richardson, A.; El-Kishky, A.; Low, A.; Helyar, A.; Madry, A.; Beutel, A.; Carney, A.; et~al. 2024.
\newblock Openai o1 system card.
\newblock \emph{arXiv preprint arXiv:2412.16720}.

\bibitem[{Kaplan et~al.(2020)Kaplan, McCandlish, Henighan, Brown, Chess, Child, Gray, Radford, Wu, and Amodei}]{kaplan2020scaling}
Kaplan, J.; McCandlish, S.; Henighan, T.; Brown, T.~B.; Chess, B.; Child, R.; Gray, S.; Radford, A.; Wu, J.; and Amodei, D. 2020.
\newblock Scaling laws for neural language models.
\newblock \emph{arXiv preprint arXiv:2001.08361}.

\bibitem[{Kim et~al.(2023)Kim, Joo, Kim, Jang, Ye, Shin, and Seo}]{kim2023cot}
Kim, S.; Joo, S.~J.; Kim, D.; Jang, J.; Ye, S.; Shin, J.; and Seo, M. 2023.
\newblock The cot collection: Improving zero-shot and few-shot learning of language models via chain-of-thought fine-tuning.
\newblock \emph{arXiv preprint arXiv:2305.14045}.

\bibitem[{Li et~al.(2024)Li, Cui, Zhao, Kong, and Bi}]{li2024gsm}
Li, Q.; Cui, L.; Zhao, X.; Kong, L.; and Bi, W. 2024.
\newblock Gsm-plus: A comprehensive benchmark for evaluating the robustness of llms as mathematical problem solvers.
\newblock \emph{arXiv preprint arXiv:2402.19255}.

\bibitem[{Li et~al.(2022)Li, Chen, Shen, Chen, Zhang, Li, Wang, Qian, Peng, Mao et~al.}]{li2022explanations}
Li, S.; Chen, J.; Shen, Y.; Chen, Z.; Zhang, X.; Li, Z.; Wang, H.; Qian, J.; Peng, B.; Mao, Y.; et~al. 2022.
\newblock Explanations from large language models make small reasoners better.
\newblock \emph{arXiv preprint arXiv:2210.06726}.

\bibitem[{Li et~al.(2025)Li, Ma, Yan, Zhang, Liu, Lu, Xu, Chen, Wang, Zhan et~al.}]{li2025model}
Li, Y.; Ma, Y.; Yan, S.; Zhang, C.; Liu, J.; Lu, J.; Xu, Z.; Chen, M.; Wang, M.; Zhan, S.; et~al. 2025.
\newblock Model Merging in Pre-training of Large Language Models.
\newblock \emph{arXiv preprint arXiv:2505.12082}.

\bibitem[{Liao et~al.(2024)Liao, He, Hao, Li, Zhang, Zhao, and Liu}]{liao2024textit}
Liao, H.; He, S.; Hao, Y.; Li, X.; Zhang, Y.; Zhao, J.; and Liu, K. 2024.
\newblock SKIntern: Internalizing Symbolic Knowledge for Distilling Better CoT Capabilities into Small Language Models.
\newblock \emph{arXiv preprint arXiv:2409.13183}.

\bibitem[{Lightman et~al.(2023)Lightman, Kosaraju, Burda, Edwards, Baker, Lee, Leike, Schulman, Sutskever, and Cobbe}]{lightman2023let}
Lightman, H.; Kosaraju, V.; Burda, Y.; Edwards, H.; Baker, B.; Lee, T.; Leike, J.; Schulman, J.; Sutskever, I.; and Cobbe, K. 2023.
\newblock Let's verify step by step.
\newblock In \emph{The Twelfth International Conference on Learning Representations}.

\bibitem[{Liu et~al.(2023)Liu, Li, Zhang, Du, Chen, Hu, Xu, Chen, and Wu}]{liu2023mind}
Liu, W.; Li, G.; Zhang, K.; Du, B.; Chen, Q.; Hu, X.; Xu, H.; Chen, J.; and Wu, J. 2023.
\newblock Mind's mirror: Distilling self-evaluation capability and comprehensive thinking from large language models.
\newblock \emph{arXiv preprint arXiv:2311.09214}.

\bibitem[{Rajbhandari et~al.(2020)Rajbhandari, Rasley, Ruwase, and He}]{rajbhandari2020zero}
Rajbhandari, S.; Rasley, J.; Ruwase, O.; and He, Y. 2020.
\newblock Zero: Memory optimizations toward training trillion parameter models.
\newblock In \emph{SC20: International Conference for High Performance Computing, Networking, Storage and Analysis}, 1--16. IEEE.

\bibitem[{Rajpurkar et~al.(2016)Rajpurkar, Zhang, Lopyrev, and Liang}]{rajpurkar2016squad}
Rajpurkar, P.; Zhang, J.; Lopyrev, K.; and Liang, P. 2016.
\newblock Squad: 100,000+ questions for machine comprehension of text.
\newblock \emph{arXiv preprint arXiv:1606.05250}.

\bibitem[{Ramos et~al.(2025)Ramos, Mamede, Jain, Canelas, Gamboa, and Le~Goues}]{ramos2025large}
Ramos, D.; Mamede, C.; Jain, K.; Canelas, P.; Gamboa, C.; and Le~Goues, C. 2025.
\newblock Are large language models memorizing bug benchmarks?
\newblock In \emph{2025 IEEE/ACM International Workshop on Large Language Models for Code (LLM4Code)}, 1--8. IEEE.

\bibitem[{Sener and Savarese(2017)}]{sener2017active}
Sener, O.; and Savarese, S. 2017.
\newblock Active learning for convolutional neural networks: A core-set approach.
\newblock \emph{arXiv preprint arXiv:1708.00489}.

\bibitem[{Setlur et~al.(2024)Setlur, Nagpal, Fisch, Geng, Eisenstein, Agarwal, Agarwal, Berant, and Kumar}]{setlur2024rewarding}
Setlur, A.; Nagpal, C.; Fisch, A.; Geng, X.; Eisenstein, J.; Agarwal, R.; Agarwal, A.; Berant, J.; and Kumar, A. 2024.
\newblock Rewarding progress: Scaling automated process verifiers for llm reasoning.
\newblock \emph{arXiv preprint arXiv:2410.08146}.

\bibitem[{Shao et~al.(2024)Shao, Wang, Zhu, Xu, Song, Bi, Zhang, Zhang, Li, Wu et~al.}]{shao2024deepseekmath}
Shao, Z.; Wang, P.; Zhu, Q.; Xu, R.; Song, J.; Bi, X.; Zhang, H.; Zhang, M.; Li, Y.; Wu, Y.; et~al. 2024.
\newblock Deepseekmath: Pushing the limits of mathematical reasoning in open language models.
\newblock \emph{arXiv preprint arXiv:2402.03300}.

\bibitem[{Srivastava et~al.(2023)Srivastava, Rastogi, Rao, Shoeb, Abid, Fisch, Brown, Santoro, Gupta, Garriga-Alonso et~al.}]{srivastava2023beyond}
Srivastava, A.; Rastogi, A.; Rao, A.; Shoeb, A.~A.; Abid, A.; Fisch, A.; Brown, A.~R.; Santoro, A.; Gupta, A.; Garriga-Alonso, A.; et~al. 2023.
\newblock Beyond the imitation game: Quantifying and extrapolating the capabilities of language models.
\newblock \emph{Transactions on machine learning research}.

\bibitem[{Suzgun et~al.(2022)Suzgun, Scales, Sch{\"a}rli, Gehrmann, Tay, Chung, Chowdhery, Le, Chi, Zhou et~al.}]{suzgun2022challenging}
Suzgun, M.; Scales, N.; Sch{\"a}rli, N.; Gehrmann, S.; Tay, Y.; Chung, H.~W.; Chowdhery, A.; Le, Q.~V.; Chi, E.~H.; Zhou, D.; et~al. 2022.
\newblock Challenging big-bench tasks and whether chain-of-thought can solve them.
\newblock \emph{arXiv preprint arXiv:2210.09261}.

\bibitem[{Touvron et~al.(2023)Touvron, Martin, Stone, Albert, Almahairi, Babaei, Bashlykov, Batra, Bhargava, Bhosale et~al.}]{touvron2023llama}
Touvron, H.; Martin, L.; Stone, K.; Albert, P.; Almahairi, A.; Babaei, Y.; Bashlykov, N.; Batra, S.; Bhargava, P.; Bhosale, S.; et~al. 2023.
\newblock Llama 2: Open foundation and fine-tuned chat models.
\newblock \emph{arXiv preprint arXiv:2307.09288}.

\bibitem[{Wang et~al.(2024)Wang, Xiong, Xie, Zhao, and Zhang}]{wang2024interpretable}
Wang, H.; Xiong, W.; Xie, T.; Zhao, H.; and Zhang, T. 2024.
\newblock Interpretable preferences via multi-objective reward modeling and mixture-of-experts.
\newblock \emph{arXiv preprint arXiv:2406.12845}.

\bibitem[{Wang et~al.(2022)Wang, Wei, Schuurmans, Le, Chi, Narang, Chowdhery, and Zhou}]{wang2022self}
Wang, X.; Wei, J.; Schuurmans, D.; Le, Q.; Chi, E.; Narang, S.; Chowdhery, A.; and Zhou, D. 2022.
\newblock Self-consistency improves chain of thought reasoning in language models.
\newblock \emph{arXiv preprint arXiv:2203.11171}.

\bibitem[{Wei et~al.(2022)Wei, Wang, Schuurmans, Bosma, Xia, Chi, Le, Zhou et~al.}]{wei2022chain}
Wei, J.; Wang, X.; Schuurmans, D.; Bosma, M.; Xia, F.; Chi, E.; Le, Q.~V.; Zhou, D.; et~al. 2022.
\newblock Chain-of-thought prompting elicits reasoning in large language models.
\newblock \emph{Advances in neural information processing systems}, 35: 24824--24837.

\bibitem[{Xu et~al.(2025)Xu, Zhou, Liu, Liu, Han, and Zhang}]{xu2025twt}
Xu, J.; Zhou, M.; Liu, W.; Liu, H.; Han, S.; and Zhang, D. 2025.
\newblock TwT: Thinking without Tokens by Habitual Reasoning Distillation with Multi-Teachers' Guidance.
\newblock \emph{arXiv preprint arXiv:2503.24198}.

\bibitem[{Ying et~al.(2024)Ying, Lin, Cao, Tang, Wang, Sun, Huang, and Yan}]{ying2024llms}
Ying, J.; Lin, M.; Cao, Y.; Tang, W.; Wang, B.; Sun, Q.; Huang, X.; and Yan, S. 2024.
\newblock Llms-as-instructors: Learning from errors toward automating model improvement.
\newblock \emph{arXiv preprint arXiv:2407.00497}.

\bibitem[{Yuan et~al.(2023)Yuan, Yuan, Li, Dong, Lu, Tan, Zhou, and Zhou}]{yuan2023scalingrelationshiplearningmathematical}
Yuan, Z.; Yuan, H.; Li, C.; Dong, G.; Lu, K.; Tan, C.; Zhou, C.; and Zhou, J. 2023.
\newblock Scaling Relationship on Learning Mathematical Reasoning with Large Language Models.
\newblock arXiv:2308.01825.

\bibitem[{Zhang et~al.(2025{\natexlab{a}})Zhang, Zhang, Zhang, Yu, Luo, and Tang}]{zhang2025cot}
Zhang, B.; Zhang, X.; Zhang, J.; Yu, J.; Luo, S.; and Tang, J. 2025{\natexlab{a}}.
\newblock Cot-based synthesizer: Enhancing llm performance through answer synthesis.
\newblock \emph{arXiv preprint arXiv:2501.01668}.

\bibitem[{Zhang et~al.(2023)Zhang, Song, Ye, and Gao}]{zhang2023towards}
Zhang, C.; Song, D.; Ye, Z.; and Gao, Y. 2023.
\newblock Towards the law of capacity gap in distilling language models.
\newblock \emph{arXiv preprint arXiv:2311.07052}.

\bibitem[{Zhang et~al.(2025{\natexlab{b}})Zhang, Wen, Wu, and Huang}]{zhang2025tinyllava}
Zhang, X.; Wen, S.; Wu, W.; and Huang, L. 2025{\natexlab{b}}.
\newblock Tinyllava-video-r1: Towards smaller lmms for video reasoning.
\newblock \emph{arXiv preprint arXiv:2504.09641}.

\bibitem[{Zhong et~al.(2023)Zhong, Cui, Guo, Liang, Lu, Wang, Saied, Chen, and Duan}]{zhong2023agieval}
Zhong, W.; Cui, R.; Guo, Y.; Liang, Y.; Lu, S.; Wang, Y.; Saied, A.; Chen, W.; and Duan, N. 2023.
\newblock Agieval: A human-centric benchmark for evaluating foundation models.
\newblock \emph{arXiv preprint arXiv:2304.06364}.

\bibitem[{Zhou et~al.(2021)Zhou, Song, Chen, Zhou, Wang, Yuan, and Zhang}]{zhou2021rethinking}
Zhou, H.; Song, L.; Chen, J.; Zhou, Y.; Wang, G.; Yuan, J.; and Zhang, Q. 2021.
\newblock Rethinking soft labels for knowledge distillation: A bias-variance tradeoff perspective.
\newblock \emph{arXiv preprint arXiv:2102.00650}.

\end{thebibliography}

\clearpage
\appendix
\setcounter{figure}{0}
\setcounter{table}{0}
\renewcommand{\thefigure}{A\arabic{figure}}
\renewcommand{\thetable}{A\arabic{table}}
\begin{center}
      {\bf APPENDIX}
    \end{center}

\section{Dataset Description}
\label{dataset_description}

\begin{table}[H]
  \centering
  \resizebox{\linewidth}{!}{
  \begin{tabular}{cccc}
    \toprule
    
    Task & Dataset & \#Train & \#Test \\
    \midrule
    \makecell{Mathematical } & GSK8K & 7473 & 1319 \\
    & GSM8KPlus & - & 3957 \\ 
    \hline
    \makecell{Complex} & BBH & 5860 & 651 \\
    \makecell{reasoning}    & AGIEval & - &  2546 \\
        & BB-sub & - & 5384 \\
    \hline
    \makecell{Factural} & MMLU & 3984 & 704 \\
    \makecell{reasoning}    & ARC-E & - &  9505 \\
        & ARC-C & - & 4689 \\
    \hline
    \makecell{Semantic matching} & MRPC & 4076 & 1725 \\

    \bottomrule
    \end{tabular}
  }
  \caption{Statistical details of the selected academic datasets}
  \label{table_dataset}
\end{table}

We list the detail of each dataset in Table~\ref{table_dataset}. 

\section{Training Hyperparameters}

We request GPT to get CoTs from teacher LLM.

\begin{itemize}[itemsep=0.5ex, parsep=0ex, topsep=0ex]
    \item Model: GPT-4-Turbo.
    \item Temperature: 1.5.
    \item Max length: 1024.
\end{itemize}

We request GPT-4-Turbo 10 times to get enough CoTs for random sampling. We list the accurate match metric of GPT model on each testset in Table~\ref{table_dataset_accuracy}. 

\begin{table}[H]
  \centering
  \resizebox{\linewidth}{!}{
  \begin{tabular}{ccc}
    \toprule
    
    Task & Dataset & Accurate Match \\
    \midrule
    \makecell{Mathematical } & GSK8K & 88.1  \\
    & GSM8KPlus & 85.2  \\ 
    \hline
    \makecell{Complex} & BBH & 69.1  \\
    \makecell{reasoning}    & AGIEval & 62.6  \\
        & BB-sub & 50.3  \\
    \hline
    \makecell{Factural} & MMLU & 77.8  \\
    \makecell{reasoning}    & ARC-E & 89.9  \\
        & ARC-C & 91.9 \\
    \hline
    \makecell{Semantic matching} & MRPC & 79.7  \\
    \bottomrule
    \end{tabular}
  }
  \caption{Accurate match of GPT models on the academic test datasets}
  \label{table_dataset_accuracy}
\end{table}

\noindent Our model is developed based on verl(Volcano Engine Reinforcement Learning for LLMs). 

\begin{itemize}[itemsep=0.5ex, parsep=0ex, topsep=0ex]
 \item Pretrained model: DeepSeek-R1-7B and Qwen2.5-7B. 
 \item Advantage estimator: GRPO.
 \item Training batch size: 160.
 \item Max input prompt length: 1024.
 \item Max output length: 1024.
 \item Actor learning rate: 1e-6.
 \item Micro batch size per GPU: 2.
 \item KL coefficient: 0.001.
 \item Rollout number: 5.
 \item Tensor parallel size: 2.
 \item Format reward: 0.1.
\end{itemize}

\section{Proof Sketch}
\label{provement}

\textbf{\textit{Objective.}} Given a set of diverse CoTs $\{C_1, C_2, \ldots, C_k\}$ sampled stochastically, we aim to show that a well-trained SLM with access to these references generates an output $C_{out}$ with higher expected reward than the CoT $C_{SLM}$ produced without such references.

Formally, let $R(C)$ denote the discrete reward of a CoT $C$. We seek to prove:
\begin{footnotesize}
\begin{equation}\label{eq:prove_1}
R(C_{\text{out}}) \geq \mathbb{E}[R(C_{SLM})].
\end{equation}
\end{footnotesize}

For simplicity, we omit the expectation over $R(C_{out})$, while it still denotes the average reasoning quality.

\textbf{\textit{Assumptions.}} To facilitate the derivation, we adopt the following assumptions:
\begin{itemize}
\item [1)]Let $e=\phi (C)$ be the semantic embedding of CoT $C$. Then, $C_{out}, C_{SLM}, C_{i}$ correspond to $e_{out}, e_{SLM}, e_{i}$. There also exists a continuous reward function:
\begin{footnotesize}
\begin{equation}\label{eq:prove_2}
\widetilde{R}(e)=\mathbb{E}_{C\sim p(C|e)}[R(C)],
\end{equation}
\end{footnotesize}
such that optimizing $\max_\theta \mathbb{E}[R(C)] $ via policy gradient is equivalent to $\max_\theta \mathbb{E}[\widetilde{R}(\phi(C))]$.

\item [2)] The reasoning ability of LLM we selected is not weaker than that of SLM when no external CoTs are provided..
\item [3)]A well-trained SLM can captures extra gains in performance beyond direct combination of inputs.
\item [4)]The sampled embeddings $e_i$ are scattered around the optimal $e^*$ in the latent space.
\end{itemize}

\textbf{\textit{Derivation.}}
According to Eq.~\eqref{eq:prove_2}, we restate the objective in Eq.~\eqref{eq:prove_1} with $\widetilde{R}(\cdot)$:
\begin{footnotesize}
\begin{equation}\label{eq:prove_3}
\widetilde{R}(e_{out}) \geq \mathbb{E}[\widetilde{R}(e_{SLM})].
\end{equation}
\end{footnotesize}

By Assumption. 2, the CoT $e_i$ sampled from the teacher LLM yields at least as high a reward as $e_{SLM}$ from the SLM without external references, that is:
\begin{footnotesize}
\begin{equation}\label{eq:prove_4}
\widetilde{R}(e_{i}) \geq \widetilde{R}(e_{SLM}).
\end{equation}
\end{footnotesize}

By substituting Eq.~\eqref{eq:prove_4} into Eq.~\eqref{eq:prove_3}, the problem reduces to proving a stricter inequality by following Eq.~\eqref{eq:prove_5}, which, if satisfied, directly implies the original objective:
\begin{footnotesize}
\begin{equation}\label{eq:prove_5}
\widetilde{R}(e_{out}) \geq \mathbb{E}_{C\sim p(C|e)}[\widetilde{R}(e_i)] =\frac{1}{k}\sum_{i=1}^{k}{\widetilde{R}(e_i)}.
\end{equation}
\end{footnotesize}

Following Assumption. 3, the SLM may either directly adopt or reject inputs from ${e_i}$, and can further improve through refinement or novel reasoning. Formally,
\begin{footnotesize}
\begin{align}\label{eq:prove_6}
\widetilde{R}(e_{out}) = \widetilde{R} \left( \sum_{i=1}^{k}w_ie_i\right) + \triangle_{gen} \geq \widetilde{R}\left(\sum_{i=1}^{k}w_ie_i\right),
\end{align}
\end{footnotesize}
where $\triangle_{gen} \geq 0$ represents additional gain from novel or refined generation, and $\sum_{i=1}^{k} w_i=1$ denotes the SLM's weighting over the $\{C_i\}$,reflecting its selective adoption or rejection of each $C_i$.

Substituting Eq.~\eqref{eq:prove_6} into Eq.~\eqref{eq:prove_5}, the remaining proof reduces to:
\begin{footnotesize}
\begin{align}
\widetilde{R}\left(\sum_{i=1}^{k}w_ie_i\right) \geq \frac{1}{k}\sum_{i=1}^{k}{\widetilde{R}(e_i)}. \label{eq:prove_7}
\end{align}
\end{footnotesize}

Following \cite{li2025model}, we prove Eq.~\eqref{eq:prove_7} in the following steps, which in turn completes the argument.

Specifically, we perform a second-order Taylor expansion of the reward function $\widetilde{R}(\cdot)$ at the optimal embedding $e^*$, where $\nabla \widetilde{R}(e^{*})=0$:
\begin{footnotesize}
\begin{align}
\widetilde{R}(e)&\approx \widetilde{R}(e^{*}) + {(e-e^{*})}^{T}\nabla \widetilde{R}(e^{*}) + \frac{1}{2}{(e-e^{*})}^{T}H(e-e^{*}) \notag \\
&= \widetilde{R}(e^{*}) + \frac{1}{2}{(e-e^{*})}^{T}H(e-e^{*}), \label{eq:prove_8}
\end{align}
\end{footnotesize}
where $H$ denotes the Hessian matrix of the reward surface at $e^*$, capturing its local curvature.

Let $\delta _i = e_i-e^*$, then applying the second-order expansion in Eq.~\eqref{eq:prove_8} to Eq.~\eqref{eq:prove_7}, the inequality becomes:
\begin{footnotesize}
\begin{align}
{(\sum_{i=1}^{k}w_i\delta_i)^T}H(\sum_{i=1}^{k}w_i\delta_i) \geq \frac{1}{k}\sum_{i=1}^{k}{{\delta_i}^{T}H\delta_i}. \label{eq:prove_9}
\end{align}
\end{footnotesize}

Expanding the left-hand side by diagonal and off-diagonal terms:
\begin{footnotesize}
\begin{align}
\sum_{i=1}^{k}w_i^2\delta_i^TH\delta_i + \sum_{i=1}^{k}\sum_{j\neq i} w_iw_j \delta_i^TH\delta_j. \label{eq:prove_10}
\end{align}
\end{footnotesize}

Assuming uniform weights $w_i = \frac{1}{k}$ $w_i \neq \frac{1}{k}$,  becomes
\begin{footnotesize}
\begin{align}
\frac{1}{k^2}(\sum_{i+1}^k\delta_i^TH\delta_i + \sum_{i=1}^{k}\sum_{j\neq i} \delta_i^TH\delta_j) \geq \frac{1}{k}\sum_{i=1}^{k}{{\delta_i}^{T}H\delta_i}.\label{eq:prove_11}
\end{align}
\end{footnotesize}

Rearranging, we aim to prove:
\begin{footnotesize}
\begin{align}
\sum_{i=1}^{k}\sum_{j\neq i} \delta_i^TH\delta_j \geq  (k-1)\sum_{i=1}^{k}\delta_i^TH\delta_i.\label{eq:prove_12}
\end{align}
\end{footnotesize}

Unlike loss functions minimized at local minima, the reward function achieves a maximum at $e^*$. Therefore, its Hessian $H$ is negative definite, implying $\delta_i^TH\delta_i \leq 0$.

Given our prior assumption that the CoTs $\{C_i\}$ are complementary, the deviation vectors $\delta_i$ and $\delta_j$ exhibit mild negative correlation—i.e., their directions point oppositely in the reward landscape. As a result, for a negative definite $H$, their cross terms $\delta_i^TH\delta_j \geq 0$. Hence:
\begin{footnotesize}
\begin{align}
\sum_{i=1}^{k}\sum_{i\neq j} \delta_i^TH\delta_j \geq 0 \geq (k-1)\sum_{i=1}^{k}\delta_i^TH\delta_i.\label{eq:prove_13}
\end{align}
\end{footnotesize}

This confirms that Eq.~\eqref{eq:prove_1} holds under uniform weighting, with equality achieved only when $\widetilde{R}(e_1) =...= \widetilde{R}(e_k) = \widetilde{R}(e^*)$.

For the more general case with non-uniform weights, a well-trained model is expected to assign higher weights to CoTs with smaller deviations $\delta_i$ (i.e., closer to optimal), and lower—even negative—weights to larger deviations. This selective weighting further increases the left-hand side of Eq.~\eqref{eq:prove_13}, making the inequality easier to satisfy. Therefore, Eq.~\eqref{eq:prove_1} is proven, that is,  Proposition~\ref{prop:reward} is proven.

In addition, Proposition~\ref{prop:decomposition} folows directly from Eq.~\eqref{eq:prove_6} and Eq.~\eqref{eq:prove_2}.

\section{Examples}
\label{examples}

\noindent \textbf{Industry dataset.} This dataset contains 210k samples, evenly distributed across three categories—irrelevant, neutral, and relevant—with a 1:1:1 ratio. Each sample is represented as a tuple $\langle query, ad\_info, label, comment \rangle$. An Example is shown in Tab.~\ref{tab:industry_example}.

Additionally, we provide a set of examples illustrating how external CoTs are generated on our industry dataset. For the data point in Tab.~\ref{tab:industry_example}, we obtain two external CoTs that lead to different answers from CoT teacher, as shown in Tab.~\ref{tab:industry_cot}. The second external CoT in Tab.~\ref{tab:industry_cot} is a rule-aware CoT, generated according to Sec.~\ref{sec:datasets} using the comment provided in Tab.~\ref{tab:industry_example}.

\begin{table}[h]
\centering

\resizebox{1.0\linewidth}{!}{
    \begin{tabular}{p{\linewidth}}
    \toprule
    \rowcolor{gray!20} \textbf{Query}\\
    How to solve the math problem about a snail climbing a tree?   \\

    \rowcolor{gray!20} \textbf{Ad\_info} \\
    This video promotes the Zuoyebang app, an educational tool that covers all the essential ancient poems children need to memorize at different learning stages. Each poem includes four key features: original recitation, text translation, literary analysis, and fun extensions. These help children understand the meaning of the poems and enhance their memorization.

    Title: Zuoyebang is here! No more worries about your child’s homework — the app can check answers online. Poor grades? Don’t worry — learning from home is easy. Parents around you are all using it — kids learn more efficiently, and parents can relax. Don’t hesitate — give it a try today! \\
    \hline
    \rowcolor{gray!20} \textbf{Label} \\
    Irrelevant \\

    \rowcolor{gray!20} \textbf{Comment} \\
    
    [Post-view Search]: Irrelevant; a typical educational query — “How to solve the snail climbing a tree problem”
    
    [User Intent]: The user is looking for a solution or explanation for the “snail climbing a tree” math problem, typically found in elementary or middle school curricula.
    
    [Assessment Process]: The retrieved ad promotes the Zuoyebang app, focusing on its features related to memorizing classical Chinese poems across grade levels, including recitation, translation, analysis, and fun extensions. It emphasizes generic benefits like “no more homework worries,” “instant answer checking,” and “learn from home,” but does not showcase any math-related content or address the specific “snail climbing a tree” problem. Given the user’s clear intent for a math solution, the ad fails to match the specific need.
    
    [Conclusion]: While the ad meets a general educational demand, it does not address math problems or the specific query, resulting in weak relevance and a tendency toward irrelevance. \\
    
    \bottomrule
    \end{tabular}
}

\caption{An example of the industry dataset: Ad Search Relevance Prediction.}
\label{tab:industry_example}
\end{table}

\textbf{GPT request example.} An example of a request to GPT is shown in Tab.~\ref{tab:GSM8KGPTExample}.

\textbf{Prompt design for SLM training.} A prompt used for training SLMs is provided in Tab.~\ref{tab:GSM8KPromptExample}.

\textbf{SLM response example.} The response generated by ReaLM is presented in Tab.~\ref{tab:GSM8KResponseExample}

\textbf{Learning from failure.} Illustrative failure cases are shown in Tab.~\ref{tab:GSM8KLearningExample-sub1} and Tab.~\ref{tab:GSM8KLearningExample-sub2}.

\begin{table*}[h]
\centering
\small
\resizebox{1.0\linewidth}{!}{
    \begin{tabular}{p{\linewidth}}
    \toprule

    \rowcolor{gray!20} \textbf{CoT Teacher's Responses} \\
    
    [CoT1] The user's query is about a math word problem involving a snail climbing a wall, with the core intent being “precise problem-solving search.” The ad promotes the Zuoyebang app, emphasizing its homework assistance features, including online answer checking, subject coverage, and problem-solving support. Although the ad highlights classical poetry resources, Zuoyebang is a comprehensive learning platform, and solving math problems is one of its core functions. The user's need for solution steps or answers is directly related to the ad’s messages like “no more homework worries” and “improve learning efficiency.” Moreover, the need for precise problem-solving aligns well with the app’s core functionality. \#\#\#\# Relevant

    [CoT2] The user's query is a math word problem — the "snail climbing a wall" problem — with a clear demand for precise problem-solving and a core need for a math solution. The ad primarily promotes Zuoyebang’s classical poetry learning features (recitation, translation, analysis, etc.) and does not mention any math-related problem-solving or search functions. Although the title includes terms like “Zuoyebang” and “online answer checking,” the video content does not cover the math domain and has no direct connection to the user’s math-related query. The ad lacks key elements such as “precise problem-solving” or “math questions,” and therefore fails to meet the user’s need. \#\#\#\# Irrelevant
    \\
    \hline
    \rowcolor{gray!20} \textbf{Label} \\
    Irrelevant \\
    \bottomrule
    \end{tabular}
}

\caption{An example of the industry dataset: Ad Search Relevance Prediction.}
\label{tab:industry_cot}
\end{table*}

\begin{table*}[t]
\centering
\small
\resizebox{1.0\linewidth}{!}{
    \begin{tabular}{p{\linewidth}}
    \toprule
    \rowcolor{gray!20} \textbf{Question}\\
    Amanda and her family are going to re-paint all the walls inside their house. Before they get started they want to divide up the work. Since all the rooms in the house have different numbers and sizes of walls in them, they figure the fairest way to divide up the work is to count all the walls in the house and assign an equal number to each person. There are 5 people in Amanda's family, including herself. There are 9 rooms in the house. 5 of the rooms have 4 walls each. The other 4 rooms each have 5 walls each. To be fair, how many walls should each person in Amanda's family paint?  \\

    \rowcolor{gray!20} \textbf{Answer} \\

    8 \\
    \hline
    \rowcolor{gray!20} \textbf{Prompt for GPT request} \\
    
    Question: Amanda and her family are going to re-paint all the walls inside their house. Before they get started they want to divide up the work. Since all the rooms in the house have different numbers and sizes of walls in them, they figure the fairest way to divide up the work is to count all the walls in the house and assign an equal number to each person. There are 5 people in Amanda's family, including herself. There are 9 rooms in the house. 5 of the rooms have 4 walls each. The other 4 rooms each have 5 walls each. To be fair, how many walls should each person in Amanda's family paint?

    Answer the question. First, you should think step by step and then answer directly after \#\#\#\# \\

    \rowcolor{gray!20} \textbf{Ground truth} \\

    8 \\

    \hline

    \rowcolor{gray!20} \textbf{GPT responses}\\

    [CoT1] Step-by-step: - Calculate the total number of walls in rooms with 4 walls: 5 rooms x 4 walls each = 20 walls. - Calculate the total number of walls in rooms with 5 walls: 4 rooms x 5 walls each = 20 walls. - Total walls in the whole house: 20 walls + 20 walls = 40 walls. - Divide the total number of walls by the number of people: 40 walls / 5 people = 8 walls. Each person in Amanda's family should paint 8 walls. \#\#\#\# 8

    [CoT2] First, find the total number of walls. Five rooms with 4 walls each is 5 × 4 = 20 walls. Four rooms with 5 walls each is 4 × 5 = 20 walls. Total walls = 20 + 20 = 40 walls. Divide 40 walls among 5 people equally: 40 ÷ 5 = 8 walls each.  Each person should paint 8 walls. \#\#\#\# 8

    [CoT3] Firstly, calculate the total number of walls in the house. Total walls = 4 + 5 = 9. \#\#\#\# 9

    [CoT4] First, calculate the total number of walls. Then, divide the total number of walls by the number of people (5) to get how many walls each person should paint.  Total walls = 4 + 5 = 9 walls. Each person paints = 9 / 5 = 1.8 walls.  \#\#\#\# 1.8
    
    [CoT5] First, determine the total number of walls. Five rooms have 4 walls each: 5 × 4 = 20 walls. Four rooms have 5 walls each: 4 × 5 = 20 walls. Total walls = 20 + 20 = 40. Dividing 40 walls by 5 people gives 8 walls each. 8 walls per person. \#\#\#\# 8
    
    [CoT6]  First, calculate the total number of walls in the house by multiplying the number of rooms with their respective number of walls and then adding those results. Then, divide the total number of walls by the number of people in Amanda's family to find out how many walls each person should paint. Each person should paint 9 walls. \#\#\#\# 9 
    
    [CoT7] First, find the total number of walls: 5 rooms * 4 walls + 4 rooms * 5 walls = 20 + 20 = 40 walls. Then divide evenly among 5 people: 40 walls ÷ 5 people = 8 walls per person. 8 walls per person. \#\#\#\# 8

    [CoT8] First, calculate total walls: 5 rooms × 4 walls = 20 walls; 4 rooms × 5 walls = 20 walls. Sum is 40 walls total. Divide 40 walls equally among 5 people: 40 ÷ 5 = 8 walls each.    Each person should paint 8 walls. \#\#\#\# 8

    [CoT9] First, calculate walls in the first room: 5 rooms × 4 walls = 20 walls. Divide 20 walls equally among 5 people: 20 ÷ 5 = 4 walls each. Then calculate walls in the second room: 4 rooms × 5 walls = 20 walls. Divide 20 walls equally among 5 people: 20 ÷ 5 = 4 walls each. Finally, add the wall: 4 walls + 4 walls = 8 walls. Each person should paint 8 walls. \#\#\#\# 8
    
    [CoT10] First, calculate the total number of walls: 5 rooms × 4 walls = 20 walls; 4 rooms × 5 walls = 20 walls. Adding them: 20 + 20 = 40 walls total. Since there are 5 people, divide 40 by 5 to get walls per person: 40 ÷ 5 = 8 walls each. \#\#\#\# 8 \\
    
    \bottomrule
    \end{tabular}
}
\caption{An example of a request to GPT on the GSM8K Dataset.}
\label{tab:GSM8KGPTExample}
\end{table*}

\begin{table*}[t]
\centering
\small
\resizebox{1.0\linewidth}{!}{
    \begin{tabular}{p{\linewidth}}
    \toprule
    \rowcolor{gray!20} \textbf{Question}\\
    Amanda and her family are going to re-paint all the walls inside their house. Before they get started they want to divide up the work. Since all the rooms in the house have different numbers and sizes of walls in them, they figure the fairest way to divide up the work is to count all the walls in the house and assign an equal number to each person. There are 5 people in Amanda's family, including herself. There are 9 rooms in the house. 5 of the rooms have 4 walls each. The other 4 rooms each have 5 walls each. To be fair, how many walls should each person in Amanda's family paint?  \\
    \hdashline
    \rowcolor{gray!20} \textbf{Answer} \\

    8 \\
    \hline
    \rowcolor{gray!20} \textbf{Prompt for RL without external CoT} \\
    Question: Amanda and her family are going to re-paint all the walls inside their house. Before they get started they want to divide up the work. Since all the rooms in the house have different numbers and sizes of walls in them, they figure the fairest way to divide up the work is to count all the walls in the house and assign an equal number to each person. There are 5 people in Amanda's family, including herself. There are 9 rooms in the house. 5 of the rooms have 4 walls each. The other 4 rooms each have 5 walls each. To be fair, how many walls should each person in Amanda's family paint?

    Think step by step and give your response after \#\#\#\#. For example \#\#\#\# 1.2 \\

    \rowcolor{gray!20} \textbf{Ground truth} \\

    8 \\

    \hline
    \rowcolor{gray!20} \textbf{Prompt for MR} \\
    Question: Amanda and her family are going to re-paint all the walls inside their house. Before they get started they want to divide up the work. Since all the rooms in the house have different numbers and sizes of walls in them, they figure the fairest way to divide up the work is to count all the walls in the house and assign an equal number to each person. There are 5 people in Amanda's family, including herself. There are 9 rooms in the house. 5 of the rooms have 4 walls each. The other 4 rooms each have 5 walls each. To be fair, how many walls should each person in Amanda's family paint?
    
    You are given the following CoTs from ChatGPT. They may right or wrong.

[CoT1] Step-by-step: - Calculate the total number of walls in rooms with 4 walls: 5 rooms x 4 walls each = 20 walls. - Calculate the total number of walls in rooms with 5 walls: 4 rooms x 5 walls each = 20 walls. - Total walls in the whole house: 20 walls + 20 walls = 40 walls. - Divide the total number of walls by the number of people: 40 walls / 5 people = 8 walls. Each person in Amanda's family should paint 8 walls. \#\#\#\# 8

[CoT2] Firstly, calculate the total number of walls in the house. Total walls = 4 + 5 = 9. \#\#\#\# 9

[CoT3] First, determine the total number of walls. Five rooms have 4 walls each: 5 × 4 = 20 walls. Four rooms have 5 walls each: 4 × 5 = 20 walls. Total walls = 20 + 20 = 40. Dividing 40 walls by 5 people gives 8 walls each. 8 walls per person. \#\#\#\# 8

[CoT4]  First, calculate the total number of walls in the house by multiplying the number of rooms with their respective number of walls and then adding those results. Then, divide the total number of walls by the number of people in Amanda's family to find out how many walls each person should paint. Each person should paint 9 walls. \#\#\#\# 9 

[CoT5] First, find the total number of walls: 5 rooms * 4 walls + 4 rooms * 5 walls = 20 + 20 = 40 walls. Then divide evenly among 5 people: 40 walls ÷ 5 people = 8 walls per person. 8 walls per person. \#\#\#\# 8

According to these CoTs, think step by step and give your response after \#\#\#\#. For example \#\#\#\# 1.2 \\

    \rowcolor{gray!20} \textbf{Ground truth} \\

    8 \\

    \hline
    \rowcolor{gray!20} \textbf{Prompt for MRPV}  \\

    Question: Amanda and her family are going to re-paint all the walls inside their house. Before they get started they want to divide up the work. Since all the rooms in the house have different numbers and sizes of walls in them, they figure the fairest way to divide up the work is to count all the walls in the house and assign an equal number to each person. There are 5 people in Amanda's family, including herself. There are 9 rooms in the house. 5 of the rooms have 4 walls each. The other 4 rooms each have 5 walls each. To be fair, how many walls should each person in Amanda's family paint?

    You are given the following CoTs from ChatGPT. They may right or wrong.

    [CoT1] Step-by-step: - Calculate the total number of walls in rooms with 4 walls: 5 rooms x 4 walls each = 20 walls. - Calculate the total number of walls in rooms with 5 walls: 4 rooms x 5 walls each = 20 walls. - Total walls in the whole house: 20 walls + 20 walls = 40 walls. - Divide the total number of walls by the number of people: 40 walls / 5 people = 8 walls. Each person in Amanda's family should paint 8 walls. \#\#\#\# 8
    
    [CoT2] Firstly, calculate the total number of walls in the house. Total walls = 4 + 5 = 9. \#\#\#\# 9
    
    [CoT3] First, determine the total number of walls. Five rooms have 4 walls each: 5 × 4 = 20 walls. Four rooms have 5 walls each: 4 × 5 = 20 walls. Total walls = 20 + 20 = 40. Dividing 40 walls by 5 people gives 8 walls each. 8 walls per person. \#\#\#\# 8
    
    [CoT4] First, calculate the total number of walls. Then, divide the total number of walls by the number of people (5) to get how many walls each person should paint.  Total walls = 4 + 5 = 9 walls. Each person paints = 9 / 5 = 1.8 walls.  \#\#\#\# 1.8
    
    [CoT5] First, calculate the total number of walls: 5 rooms × 4 walls = 20 walls; 4 rooms × 5 walls = 20 walls. Adding them: 20 + 20 = 40 walls total. Since there are 5 people, divide 40 by 5 to get walls per person: 40 ÷ 5 = 8 walls each. \#\#\#\# 8
    
    Think step by step. You should firstly judge the correctness of each cot, and then give your response after \#\#\#\#. For example \#\#\#\# right right right wrong wrong 1.2 \\

    \rowcolor{gray!20} \textbf{Ground truth} \\

    right wrong right wrong right 8 \\
    
    \bottomrule
    \end{tabular}
}
\caption{A prompt used for MR and MRPV on the GSM8K Dataset.}
\label{tab:GSM8KPromptExample}
\end{table*}

\begin{table*}[t]
\centering
\small
\resizebox{1.0\linewidth}{!}{
    \begin{tabular}{p{\linewidth}}
    \toprule
    \rowcolor{gray!20} \textbf{Question}\\
    Courtney attended a concert and reported that the audience was 48 in number. However, Kelly went to the same concert and said that Courtney had made the mistake of overstating the number of people in attendance by 20\%. If Kelly was right, how many people really attended the concert?  \\
    \rowcolor{gray!20} \textbf{Answer} \\

    40 \\

    \hline
    
    \rowcolor{gray!20} \textbf{Prompt for MRPV}  \\

    Question: Courtney attended a concert and reported that the audience was 48 in number. However, Kelly went to the same concert and said that Courtney had made the mistake of overstating the number of people in attendance by 20\%. If Kelly was right, how many people really attended the concert?.
    
    [CoT1] Courtney stated there were 48 people. Kelly claims Courtney overstated by 20\%, meaning 48 represents 120\% of the actual number. To find the real attendance, divide 48 by 1.2 (since 120\% = 1.2). This gives the actual number of attendees. \#\#\#\# 40
    
    [CoT2] Courtney reported 48 people at the concert. Kelly said this was 20\% more than the actual number. That means 48 is 120\% of the real attendance. To find the actual attendance, divide 48 by 1.2. \#\#\#\# 40
    
    [CoT3] First, consider that Courtney reported 48 people, which Kelly claims is a 20\% overestimate. This means the reported number is 120\% (or 1.2 times) of the actual attendance. To find the real number of people, divide 48 by 1.2. \#\#\#\# 40
    
    [CoT4] Courtney reported 48 attendees. According to Kelly, 48 is an overestimate by 20\%, meaning it's 120\% of the actual number. To find the real attendance, divide 48 by 1.20, as 1.20 represents the 120\% (100\% actual + 20\% overstatement). \#\#\#\# 40
    
    [CoT5] Step 1: Understand the information - Courtney reports 48 people at a concert. Step 2: Identify Kelly's report - Kelly claims that amount was overstated by 20\%. Step 3: Calculate the real number - take 20\% of Courtney's report, then subtract from it. Now: 20\% of 48 = 9.6, so real attendance according to Kelly = 48 - 9.6. There were actually 38.4 people who attended the concert, which can be rounded to 38 people when accounting for whole attendees. \#\#\#\# 38.4
    
    Your should firstly judge the correctness of each cot and then give your response after \#\#\#\#. For example <answer> \#\#\#\# right wrong right wrong wrong 0.5 \\

    \rowcolor{gray!20} \textbf{Ground truth} \\

    right right right right wrong 40 \\

    \hline
    \rowcolor{gray!20} \textbf{DeepSeek-R1-7B response} \\

    <think> Alright, let\'s break this down step by step. Courtney said there were 48 people at the concert. Kelly thinks that\'s 20\% too high. So, if 48 is 20\% more than the actual number, I need to find what the real number is.
    
    First, I\'ll represent the actual number of attendees as "x". Since 48 is 20\% more than x, that means 48 is equal to x plus 20\% of x. In equation form, that\'s 48 = x + 0.2x, which simplifies to 48 = 1.2x.
    
    To find x, I\'ll divide both sides by 1.2. So, x = 48 / 1.2. Calculating that gives x = 40. 
    
    I should check each of the claims to make sure they\'re correct. 
    
    The first four points (CoT1 to CoT4) all agree that dividing 48 by 1.2 gives 40, which seems right. The fifth point mentions subtracting 20\% of 48, which seems not equal to over counting by 20\%. So I\'ll mark CoT1 to CoT4 as right and CoT5 as wrong.</think>
    
    <answer> \#\#\#\# right right right right wrong 40 \\
    
    \bottomrule
    \end{tabular}
}
\caption{A response example from ReaLM initialled by DeepSeek-R1-7B on GSM8K Dataset}
\label{tab:GSM8KResponseExample}
\end{table*}

\begin{table*}[t]
\centering
\small
\resizebox{1.0\linewidth}{!}{
    \begin{tabular}{p{\linewidth}}
    \toprule
    \rowcolor{gray!20} \textbf{Trainining data case} \\

    John had \$200. He gave 3/8 of his money to his mother and 3/10 to his father. How much money did John have left?
    
    [CoT1]: First, calculate the amount given to his mother: \$200 * 3/8 = \$75. Next, calculate what was given to his father: \$200 * 3/10 = \$60. Total money given away is \$75 + \$60 = \$135. Total money left: initial money - given money, which equals \$200 - \$135. John has \$65 left. \#\#\#\# 65 
    
    [CoT2]: First calculate the fraction given to mother: 3/8 of \$200 = \$75. Next, calculate the fraction given to father: 3/10 of \$200 = \$60. The money given away totals \$75 + \$60 = \$135. John had \$200, substract the amount given away: \$200 - \$135 = \$65 left. John had \$65 left. \#\#\#\# 65
    
    [CoT3]: John had \$200. He gave 3/8 to his mother, which equates to \$200*(3/8) = \$75. He gave another 3/10 to his father, which is \$200*(3/10) = \$60. The total spent is \$75 (mother) + \$60 (father) = \$135. Consequently, he had \$200 (initial) - \$135 (spent) remaining. John had \$65 left. \#\#\#\# 65
    
    [CoT4]: First, find how much money John gave to his mother: (3/8) × 200 = 75. Then, calculate how much he gave his father: (3/10) × 200 = 60. Add these amounts: 75 + 60 = 135. Subtract from the total: 200 - 135 = 65. John had \$65 left. \#\#\#\# 65
    
    [CoT5]: The amount given to his mother is \$200 * 3/8 = \$75. The amount given to his father is \$200 * 3/10 = \$60. Total money given away is \$75 + \$60 = \$135. \#\#\#\# 135

    [For MR] According to these CoTs, think step by step and give your response after \#\#\#\#. For example \#\#\#\# 1.2 

    [For MRPV] Your should firstly judge the correctness of each cot and then give your response after \#\#\#\#. For example <answer> \#\#\#\# right wrong right wrong wrong 0.5 \\

    \rowcolor{gray!20} \textbf{Ground truth}\\
    right right right right wrong 65 \\

    \rowcolor{gray!20} \textbf{MRPV training output}\\

    <think> Alright, let's break down John's money situation step by step. He starts with \$200. 
    
    First, he gives 3/8 of his money to his mother. To find out how much that is, I'll calculate 3/8 of 200. So, 3/8 of 200 is 75. So, he gives \$75 to his mother.
    
    Next, he gives 3/10 of his money to his father. Again, I'll calculate 3/10 of 200. That's 60. So, he gives \$60 to his father.
    
    Now, I need to find out how much he has left after giving money to both his mother and father. He started with \$200, gave away \$75, and then \$60. So, the total given away is 75 + 60, which equals \$135.
    
    Subtracting the total given away from the initial amount: 200 - 135 equals \$65. Therefore, John has \$65 left.
    
    Cot1, CoT2, CoT3, CoT4 all agree with 65, which seems right. The fifth point mentions \$135, which seems equal to the total given away. So CoT5 is not right. </think>
    
    <answer> \#\#\#\# right right right right wrong 65 \\

    \hline

    \rowcolor{gray!20} \textbf{Similar test data}\\

    Question: Maggie spent a quarter of her money, while Riza spent one-third of her money. They each had \$60. How much money do the two of them have left?
    
    [CoT1] First, calculate the amount spent by Maggie: 1/4 of \$60, then calculate Riza's: 1/3 of \$60. Subtract both amounts from their initial sums to find out how much each has left. Add their remaining amounts together for the total they collectively have left. Maggie spent \$15, left with \$45. Riza spent \$20, left with \$40. Maggie and Riza together have \$85 left. \#\#\#\# 85
    
    [CoT2] Step-by-step: Maggie spent 1/4 of her money, so she has 3/4 remaining. Riza spent 1/3 of her money, so she has 2/3 remaining. The question mentions that they each had \$60 originally. Maggie's remaining money: (3/4) × 60 = 45 Riza's remaining money: (2/3) × 60 = 40 Total left = 45 + 40 = 85 They have \$85 left in total. \#\#\#\# 85

    [CoT3] Maggie had an initial amount, then spent a quarter, so she kept 3/4. Riza had an initial amount, spent one-third, kept 2/3. After spending, each had \$60 left. Sum total left = 60 + 60 = 120. \#\#\#\# 120
    
    [CoT4] Maggie initially has \$60 and spends 0.25*60 = \$15. She has \$60 - \$15 = \$45 left. Riza has \$60 and spends 0.33*60 \$20. She has \$60 - \$20 = \$40 left. Combined, Maggie has \$45 and Riza \$40. Maggie and Riza have \$45 + \$40 = \$85 left together. \#\#\#\# 85
    
    [CoT5] To find the answer: 1. Calculate how much money Maggie and Riza each spent. - Maggie spent 1/4 of \$60 = \$15. - Riza spent 1/3 of \$60 = \$20. 2. Determine how much money they each have left. - Maggie has \$60 - \$15 = \$45. - Riza has \$60 - \$20 = \$40. 3. Add their remaining amounts together to get the total. - Total = \$45 + \$40. Maggie and Riza have a total of \$85 left. \#\#\#\# 85
    
    [For MR] According to these CoTs, think step by step and give your response after \#\#\#\#. For example \#\#\#\# 1.2 

    [For MRPV] Your should firstly judge the correctness of each cot and then give your response after \#\#\#\#. For example <answer> \#\#\#\# right wrong right wrong wrong 0.5 \\

    \rowcolor{gray!20} \textbf{Ground truth} \\

    right right wrong right right 85 \\
    
    \bottomrule
    \end{tabular}
}
\caption{Learning from Failure: A Case from the GSM8K Dataset (Part 1)}
\label{tab:GSM8KLearningExample-sub1}
\end{table*}

\begin{table*}[t]
\centering
\small
\resizebox{1.0\linewidth}{!}{
    \begin{tabular}{p{\linewidth}}
    \toprule

    \rowcolor{gray!20} \textbf{Response from MR} \\

    <think> Maggie spent a quarter of her money, and Riza spent one-third of her money. After spending, both of them ended up with \$60 each. The question is asking how much money they have left together.
    
    Let me break it down step by step. First, let's consider Maggie. She spent a quarter of her money. Let's denote her initial amount as M. She spent (1/4)M, so the remaining amount is M - (1/4)M = (3/4)M. According to the problem, this remaining amount is \$60. So, (3/4)M = \$60. To find M, I can solve for it:(3/4)M = 60  Multiply both sides by (4/3):  M = 60 * (4/3)  M = 80. So, Maggie initially had \$80. She spent a quarter, which is \$20, so she has \$60 left. That checks out.
    
    Now, Riza. She spent one-third of her money. Let's denote her initial amount as R. She spent (1/3)R, so the remaining amount is R - (1/3)R = (2/3)R. The problem states that this remaining amount is \$60. So,(2/3)R = 60. Multiply both sides by (3/2): R = 60 * (3/2) R = 90. So, Riza initially had \$90. She spent one-third, which is \$30, so she has \$60 left. That also checks out.
    
    Now, to find the total money they have left together, we add the remaining amounts: Maggie: \$60  Riza: \$60. Total: \$60 + \$60 = \$120. Yes, that makes sense. So the correct answer should be \$120.
    
    The two of them have \$120 left together. </think>
    
    <answer>\#\#\#\# 120 \\

    \rowcolor{gray!20} \textbf{Response from MRPV} \\

    <think> Okay, so I need to figure out how much money Maggie and Riza have left after spending a portion of their money. Let me go through each of the provided COTs to see if they're correct.
    
    Starting with [CoT1]: It says Maggie spent a quarter of her money, which is 1/4 of \$60. That's \$15, so she has \$45 left. Riza spent one-third of her \$60, which is \$20, so she has \$40 left. Adding \$45 and \$40 gives \$85. That seems correct.
    
    [CoT2] talks about Maggie spending \$15 and having \$45, and Riza spending \$20 and having \$40. Adding those gives \$85. That's the same as in the first COT, so it's correct too.
    
    [CoT3] breaks it down similarly. Maggie's spending is \$15, leaving her with \$45. Riza's spending is \$20, leaving her with \$40. Total is \$85. Correct.
    
    [CoT4] is a bit more detailed. It calculates the amounts spent and subtracts from \$60 for each. It arrives at \$45 and \$40, adding to \$85. That's correct.
    
    [CoT5]: It add the original \$60 directly, but not calculate each remaining money. That's not correct. </think>
    
    <answer> \#\#\#\# right right right right wrong 85\\
    
    \bottomrule
    \end{tabular}
}
\caption{Learning from Failure: A Case from the GSM8K Dataset (Part 2)}
\label{tab:GSM8KLearningExample-sub2}
\end{table*}

\clearpage

\end{document}